\documentclass[11pt,twocolumn]{article}
\usepackage{apalike}

\usepackage{amsmath}

\usepackage{algorithm}
\usepackage{algorithmic}

\usepackage{color}

\usepackage{graphicx}

\usepackage{multirow}
\usepackage{xcolor,colortbl}

\newcommand{\rmax}{{\textsc{R-max}}}
\newcommand{\LL}{{\cal L}}

\newcommand{\TT}{{\cal T}}

\title{Active Learning for Autonomous Intelligent Agents:\\ Exploration, Curiosity, and Interaction}

\author{Manuel Lopes\\Inria Bordeaux Sud-Ouest, France\\ \textit{manuel.lopes@inria.fr} \and Luis Montesano\\University of Zaragoza, Spain\\ \textit{luis.montesano@unizar.es}}

\begin{document}

\maketitle
\pagestyle{plain}

\begin{abstract}
In this survey we present different approaches that allow an intelligent agent to explore autonomous its environment to gather information and learn multiple tasks. Different communities proposed different solutions, that are in many cases, similar and/or complementary. These solutions include active learning, exploration/exploitation, online-learning and social learning. The common aspect of all these approaches is that it is the agent to selects and decides what information to gather next. Applications for these approaches already include tutoring systems, autonomous grasping learning, navigation and mapping and human-robot interaction. We discuss how these approaches are related, explaining their similarities and their differences in terms of problem assumptions and metrics of success. We consider that such an integrated discussion will improve inter-disciplinary research and applications.\footnote{Draft v0.7 18Dec2013}
\end{abstract}

\section{Introduction}
\label{sec:intro}


One of the most remarkable aspects of human intelligence is its adaptation to new situations, new tasks and new environments. To fulfill the dream of Artificial Intelligence and to build truly Autonomous Intelligent Agents (Robots included), it is necessary to develop systems that can adapt to new situations by learning fast how to behave or how to modify their previous knowledge. Consequently, learning has taken an important role in the development of such systems. This paradigm shift has been motivated by the limitations of other approaches to cope with complex open-ended problems and fostered by the progress achieved in the fields of statistics and machine learning. 
Since tasks to be learned are becoming increasingly complex, have to be executed in ever changing environments and may involve interactions with people or other agents, learning agents are faced with situations that require either a lot of data to model and cover high dimensional spaces and/or a continuous acquisition of new information to adapt to novel situations. 
Unfortunately, data is not always easy and cheap, but often requires a lot of time, energy, computational or human resources and can be argued to be a limiting factor in the deployment of systems where learning is a key factor.

Consider for instance a robot learning from data obtained during operation. 
It is common to decouple the acquisition of training data from the learning process. However, the embodiment in this type of systems provides a unique opportunity to exploit an active learning (AL) \footnote{Active learning can also be used to describe situations where the student is involved in the learning process as opposed to passively listening to lectures, see for instance \cite{linder2001facilitating}.} approach (AL)\cite{Angluin88queries,thrun1995exploration,Settlesactivelearsurv} to guide the robot actions towards a more efficient learning and adaptation and, consequently, to achieve a better performance more rapidly.  

The robot example illustrates the main particularity of learning for autonomous agents: the abstract learning machine is embodied in a (cyber) physical environment and so it needs to find the relevant information for the task at hand by itself.
Although these ideas have been around for more than twenty years \cite{Schmidhuber91curiosity,thrun1992efficient,dorigo1994robot,aloimonos1988active}, in the last decade there has been a renewed interest from different perspectives in actively gathering data during autonomous learning. 
%
%
%
%
%
Broadly speaking, the idea of AL is to use the current knowledge the system has about the task that is currently being learned to select the most informative data to sample. 
In the field of machine learning this idea has been envigorated by the existence of huge amounts of unlabeled data freely available on the internet or from other sources. Labeling such data is expensive as it requires the use of experts or costly procedures. If similar accuracy can be obtained with less labeled data then huge savings, monetary and/or computational, could be made. 

In the context of intelligent system, another line of motivation and inspiration comes from the field of \textit{artificial development} \cite{Schmidhuber91curiosity,Weng01Science,asadaetal01cognitivedevel,lungarella03develsurvey,oudeyerEncyclo11}. This field, inspired by developmental psychology, tries to understand biological development by creating computational models of the process that biological agents go through their lifetimes. In such process there is no clearly defined tasks and the agents have to create their own representations, decide what to learn and create their own learning experiments.

A limiting factor on active approaches is the limited theoretical understanding of some of its processes. Most theoretical results on AL are recent \cite{Settlesactivelearsurv,dasgupta05greedyact,dasgupta2011two,nowak2011geometry}. The first intuition on why AL might required a smaller number of labeled data, is to note that the system will only ask for data that might changes its hypothesis and so uninformative examples will not be used. Nevertheless, previous research provides an optimistic perspective on the applicability of AL for real application, and indeed there already many examples: image classification \cite{qi2008two}, text classification \cite{Tong01}, multimedia \cite{wang2011active}, among many others (see \cite{Settlesactivelearsurv} for a review). Active learning can also be used to plan experiments in genetics research, e.g. the robot scientist \cite{king2004functional} eliminates redundant experiments based on inductive logic programming. Also, most algorithms already have an active extension: logistic regression \cite{Schein07}, support vector machines \cite{Tong01}, GP \cite{Kapoor07}, neural networks \cite{Cohn96}, mixture models \cite{Cohn96}, inverse reinforcement learning \cite{macl09airl}, among many others.
%
%

In this paper we take a very broad perspective on the meaning of AL: any situation where an agent (or a team) \textit{actively} looks for data instead of passively waiting to receive it. 
The previous description rules out those cases where a learning process uses data previously obtained in any possible way (e.g. by random movements, or with a predefined paths; or by receiving data from people or other agents). 
Thus, the key property of such algorithms is the involvement of the agent to decide what information suits better its learning task. 
There are multiple intances of this wide definition of AL with sometimes unexplored links. We structured them in three big groups: a) exploration where an agent explores its environment to learn; b) curiosity where the agent discovers and creates its own goals; and c) interaction where the existence of a human-in-the-loop is taken explicitly into account.

\begin{table}[!htbp]
\caption{Glossary}
{\small
\fbox{\parbox{0.95\columnwidth}{
\textbf{Classical Active Learning}(AL), refers to a set of approaches in which a learning algorithm is able to interactively query a source of information to obtain the desired outputs at new data points \cite{Settlesactivelearsurv}.

\textbf{Optimal Experimental Design}(OED), an early perspective on active learning where the design of the experiments is optimal according to some statistical criteria \cite{Schonlau98}. Usually not considering the interactive perspective of sensing.

\textbf{Learning Problem}, refers to the problem of estimating a function, including a policy, from data. The measures of success for such a problem vary depending on the domain. Also known as the pure exploration problem.

\textbf{Optimization Problem}, refers to the problem of finding a particular value of an unknown function from data. When compared with the Learning Problem, it is not interested in estimating the whole unknown function.

\textbf{Optimization Algorithm}, refers to methods to find the maximum/minimum of a given function. The solution to this problem might require, or not, learn a model of the function to guide exploration. We distinguish it from the Learning Problem due to the its specificities.

\textbf{Bayesian Optimization}, class of methods to solve an optimization problem that use statistical measures of uncertainty about the target function to guide exploration \cite{brochu2010tutorial}.

\textbf{Optimal Policy}, in the formalism of markov decision process, the optimal policy is the policy that provides the maximum expected (delayed) reward. We will use it also to refer to any policy, exploration or not, that is optimal according to some criteria.

\textbf{Exploration Policy}, defines the decision algorithm, or policy, that selects which actions are selected during the active learning process. This policy is not, in general, the same as the optimal policy for the learning problem. See a discussion at \cite{duff2003design,csimcsek2006intrinsic,golovin10adaptive,Toussaint12BayesianSearchGame}.

\textbf{Empirical Measures}, class of measures that estimate the progress of learning by measuring empirically how recent data as allowed the learning task to improve.
}
}}
\end{table}

\subsection{Exploration}
Exploration by an agent (or a team of agents) is at the core of rover missions, search and rescue operations, environmental monitoring, surveillance and security, best teaching strategies, online publicity, among others. 
In all these situations the amount of time and resources for completing a task is limited or unknown. Also, there are often trade-offs to be made between different tasks such as surviving in a hostile environment, communicating with other agents, gathering more information to minimize risk, collecting and analyzing samples.
All these tasks must be accomplished in the end but the order is relevant inasmuch as is it helps subsequent tasks. For instance collecting geological samples for analysis and communicating the results will be easier if the robot has already a map of the environment. Active strategies are of paramount importance to select the right tasks and actively execute the task maximizing the operation utility while minimizing the required resources or the time to accomplish the goal. 
\subsection{Curiosity}
A more open-ended perspective on learning should consider cases where the task itself is not defined. Humans develop and grow in an open-ended environment without pre-defined goals. 
Due to this uncertainty we cannot assume that all situations are considered a-priori and the agent itself has to adapt and learn new tasks. Even more problematic is that the tasks faced are so complex that learning them might require the acquisition of new skills. 

Recent results from neuroscience have given several insights into visual attention and general information seeking in humans and other animals. Results seem to indicate that curiosity is an intrinsic drive in most animals \cite{Gottlieb13TICS}.
Similarly to animals with complex behaviors, an initial period of immaturity dedicated to play and learning might allow to develop such skills. This is the main idea of developmental robotics \cite{Weng01Science,asadaetal01cognitivedevel,elman1997rethinking,lungarella03develsurvey,oudeyerEncyclo11} where the complexity of the problems that the agent is able to solve increases with time. During this period the agent is not solving a task but learning for the sake of learning. This early stage is guided by curiosity and intrinsic motivation \cite{barto2004intrinsically,Schmidhuber91curiosity,oudeyer2005playground,Singh2005intrinsically,Schmidhuber06,Oudeyer2007} and its justification is that it is a skill that will lead to a better adaptation to a large distribution of problems \cite{Singh2010}. 

\subsection{Interaction}

Learning agents have intensively tackled the problem of acquiring robust and adaptable skills and behaviors for complex tasks from two different perspectives: programming by demonstration (a.k.a. imitation learning) and learning through experience. 
From an AL perspective, the main difference between these two approaches is the source of the new data. Programming by demonstration is based on examples provided by some external agent (usually a human). Learning through experience exploits the embodiment of the agent to gather examples by itself by acting on the world. 
In the abstract AL from machine learning the new data/labels used to come from an oracle and no special regard is given to what exactly the oracle is besides well behaved properties such as no bias and consistency. More recently, data and labels may come from ratings and tagging provided by humans resulting in bias and inconsistencies. This is also the case for agents interacting with humans in which applications had taken into account where that information comes from and what other sources of information might be exploited. For instance, sometimes humans may provide more easily information other than labels that can further guide exploration\footnote{I don't like the last sentence with the last changes in the section}.

\subsection{Organization}
\label{sec:Organization}


This review will consider AL in this general setting. We will first clarify the AL principles for autonomous intelligent agents in Sec. \ref{sec:ActiveLearningFramework}. Then the core review will be organized in three main parts: Sec \ref{sec:selfguidedexplor} AL during self-exploration; Sec. \ref{sec:Curiosity} autonomous discovery/creation of goals; and finally Sec \ref{sec:explorsocial} AL with humans.
%
%

\section{Active Learning for Autonomous Intelligent Agents}
\label{sec:ActiveLearningFramework}

\begin{figure*}[htbp]
	\centering
		\includegraphics[width=0.5\columnwidth]{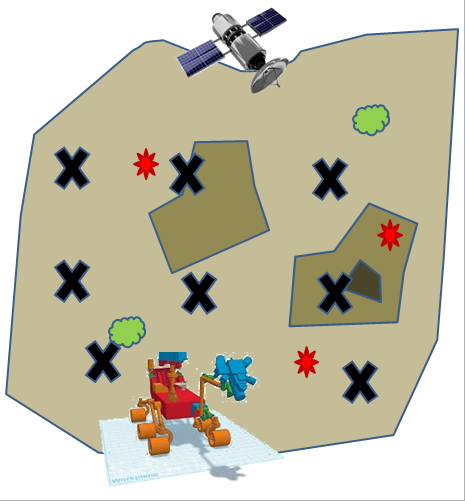}\hfill
		\includegraphics[width=0.5\columnwidth]{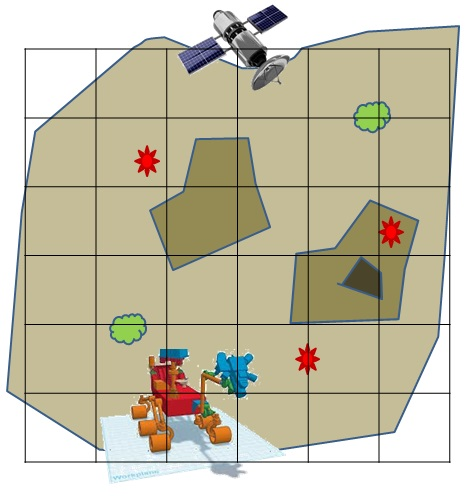}\hfill
		\includegraphics[width=0.5\columnwidth]{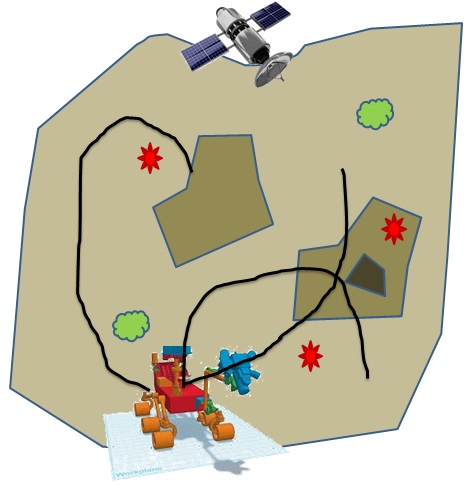}
	\caption{Different choices on active learning. A robot might choose: to look for the most informative set of sampling locations, ignoring the travel and data acquisition cost and the information gathered on the way there, either by selecting a) among an infinite set of location  or  b) by reducing its choices to a pre-defined set of locations; or c) consider the best path including the cost and the information on the way.}
	\label{fig:3perspectives}
\end{figure*}

In this Section we provide an integrated perspective on the many approaches for active learning. The name active learning has mostly been used in machine learning but here we consider any situation where a learning agent uses its current hypothesis about the learning task to select what/where/how to learn next. Different communities formulated problems with similar ideas and all of them can be useful for \textit{autonomous intelligent agents}. Different approaches are able to reduce the time, or samples, required to learn but they consider different fitness functions, learning algorithms and choices of what can be selected. Figure \ref{fig:3perspectives} shows the three main perspectives on for single task active learning. Exploration in reinforcement learning \cite{suttonbarto98}, bayesian optimization \cite{brochu2010tutorial}, multi-armed bandits \cite{bubeck2012MAB}, curiosity \cite{oudeyer2007intrinsic}, interactive machine learning \cite{Breazeal2004} or active learning for classification and regression problems \cite{Settlesactivelearsurv}, all these share many properties and face similar challenges. Interestingly, a better understanding of the different approaches from the various communities can lead to more powerful algorithms. Also, in some cases to solve the problem of one community, it is necessary to rely on the formalism of another. For instance, active learning for regression methods that can synthesize queries need to find the most informative point. This is, in general, an optimization problem in high-dimension and it is not possible to solve it exactly. Bayesian optimization methods can then be used to find the best point with a minimum of function evaluations \cite{brochu2010tutorial}. Another example, still for regression, is to decompose complex regression functions to a set of local regressions and then rely on multi-armed bandit algorithms to balance exploration in a more efficient way \cite{maillard2012hierarchical}.

\begin{figure}[htbp]
	\centering
		\includegraphics[width=0.7\columnwidth]{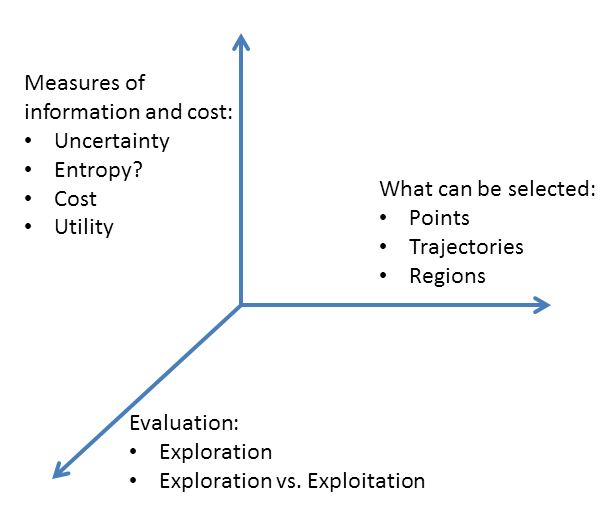}
	\caption{During autonomous exploration there are different choices that are made by an intelligent agent. These include what does the agent selects to explore; how does it evaluate its success; and how does it estimate the information gain of each choice.}
	\label{fig:axisofactive}
\end{figure}

Each of these topics would benefit from a dedicated survey and we do not aim at a definite discussion on all the methods. In this section we will discuss all these approaches with the goal of understanding the similarities, strengths and domains of application. Due to the large variety of methods and formalism we can not describe the full details and mathematical theory but we will provide references for most methods. This Section can be seen as a cookbook of active learning methods where all the design choices and tradeoffs are explained jointly with links for the theory and for examples of application (see Figure \ref{fig:axisofactive} for a summary).

\subsection{Optimal Exploration Problem}
\label{sec:OptimalExplorationProblem}

To ground the discussion, let us consider a robot whose mission is to build a map of some physical quantities of interest over a region (e.g. obstacles, air pollution, density of traffic, presence of diamonds...).
The agent will have a set of on-board capabilities for acting in the environment that will include moving along a path or to a specific position and using its sensors to obtain measurements about the quantity of interest. In addition to this, it may be possible to make decisions about other issues such as what algorithms should be used to process the obtained measurements or to fit the model of the environment. The set of all possible decisions will define the space of exploration policies\footnote{The concept is similar to the policy for reinforcement learning, but here the policy is not optimizing total reward but, instead, exploration gain (to be defined latter)} $\Pi$. 
To derive an active algorithm for this task, we need to model the costs and the loss function associated to the actions of a specific exploration policy $\pi$. The most common costs include the cost of using each of the on-board sensors (e.g. energy consumption, time required to acquire the measurement or changes in the payload) and the cost of moving from one location to another (e.g. energy and the associated autonomy constraints).  
Regarding the loss function, it has to capture the error of the learned model w.r.t. the unknown true model. For instance, one may consider the uncertainty of the predictions at each point or the uncertainty on the locations of the objects of interest.

The optimal exploration policy is the one that simultaneously gives the best learned model but with the smallest possible cost:
\begin{equation}
\pi^*_{t} = \arg max_{\pi \in \Pi} f(\pi, C(\pi), \LL_{x\in \mathcal{X}}(\hat{g}(x; D \sim \pi))),
\label{eq:costopt}
\end{equation}
where $\pi$ is an exploration policy (i.e. a sequence of actions possibly conditioned on the history of states and/or actions taken by the agent), $\Pi$ denotes the space of possible policies, $f()$ is a function that summarizes the utility of the policy\footnote{Note that $\pi$ might have different semantics depend on the task at hand. It can be an exploration policy used to learn a model $g()$ in a pure learning problem, or it can be an exploitation policy in an optimization setting. For a more detailed description on the relation of the exploration policy with the learning task see \cite{duff2003design,csimcsek2006intrinsic,golovin10adaptive,Toussaint12BayesianSearchGame}.}, and $\mathcal{X}$ is a space of points that can be sampled. Function $f$ depends on the policy itself, the cost $C(\pi)$ of executing this policy and the loss of the policy $\LL_x(\hat{g}(x; D \sim \pi))$.  The loss depends on a function $\hat{g}()$ learned with the dataset $D$ acquired following policy $\pi$. Equation \ref{eq:costopt} selects the best way to act, taking into account the task uncertainty along time. Clearly this problem is, in general, intractable and the following sections describe particular instantiations, approximations and models of this \textit{optimal exploration problem} \cite{csimcsek2006intrinsic}.

Equation \ref{eq:costopt} is intentionally vague with respect to several crucial aspects of the optimization process. For instance, time is not included in any way, and just the abstract policy $\pi$ and the corresponding policy space $\Pi$ are explicit. Also, many different costs and loss models can be fed into the function $f()$, with the different choices resulting in different problems and algorithms. 
It is the aim of this work to build  bridges between this general formulation and the problems and solutions proposed in different fields. However, before delving into the different instances of this problem, we briefly describe the three most common frameworks for active learning and then discuss possible choices for the policy space $\Pi$ and the role of the terms $C$ and $L$ in the context of autonomous agents.  
\subsection{Learning Setups}
\label{sec:TypicalProblems}

\subsubsection{Function approximation}
\label{sec:Regression}

Regression, and classification, problems are the most common problems in machine learning methods. In both cases given a dataset of points $D=\{(x,y)\}$ the goal is to find an approximation of the input output relation $g:x\rightarrow y$. Typical loss functions are the squared mean error $L~=~|\hat{h}-y|^2=|g(x)-y|^2$ for regression and the $0-1$ loss $L_{0-1}=\mathcal{I}(\hat{g}(x)=y)$ for classification, with $\mathcal{I}$ denoting the indicator function.
In this setup the cost function directly measures the cost of obtaining measurements (e.g. collecting the measurement or moving to the next spot), if it exists. The active learning perspective corresponds to deciding for which input $x$ it is more relevant to ask for the corresponding label $y$. Some other restrictions can be included such as being restricted to a finite set of input points (pool-based active learning) or having the points arriving sequentially and having to decide to query or not (online learning)(see \cite{Settlesactivelearsurv} for a comprehensive discussion on the different settings).


\subsubsection{Multi-Armed Bandits}
\label{sec:MultiArmedBandits}

An alternative formalism that is usually applied to discrete selection problems is the \textit{multi-armed bandit} (MAB) formalism \cite{gittins1979bandit,bubeck2012MAB}. Multi-arm bandits define a problem where a player, at each round, can choose an arm among a set of $n$ possible ones. After playing the selected arm the player receives a reward. In the most common setting the goal of the player is to find a strategy that allows it to get the maximum possible cumulative reward. The loss in bandit problems is usually based on the concept of regret, that is, the difference between the reward that was collected and the reward that would have been collect if the player knew which was the best arm since the beginning \cite{auer2003nonstochastic}. Many algorithms have been proposed for different variants of the problems where instead of regret the player is tested after a learning period and it has either to declare what is the best arm \cite{gabillon2011MABident} or the value of all the arms \cite{Carpentier11ActMAB}.

\subsubsection{MDP}
\label{sec:MDP}

The most general, and well known, formalism to model sequential decision processes are \textit{markov-decision process} (MDP)\cite{bellman1952theory}. When there is no knowledge about the model of the environment and an agent has to optimize a reward function while interacting with the environment the problem is called reinforcement learning (RL) \cite{suttonbarto98}. A sequential problem is modeled as a set of states $S$, actions that allow the system to change between state $A$ and the rewards that the system receives at each time step $R$. The time evolution of the system is considered to depend on the current state $s_t$ and the chosen action $a_t$, i.e. $p(s_{t+1}|s_t,a_t)$. The goal of the agent is to find a policy, i.e. $\pi(s,a)=p(a|s)$, that maximizes the total discounted reward $J(s_0)=\sum_{t=0}^\infty \gamma^t r_t$. For a complete treatment on the topic refer to \cite{Kaelbling96jmlr,suttonbarto98,szepesvari2011reinforcement,kober2013reinforcement}. 
As the agent does not know the dynamics and the reward function it can not act optimally with respect to the cost function without first exploring the environment for that information. Then it can explicitly create a model of the environment and exploit it \cite{Hester2011survey,NguyenTuong11modellearning} and directly try to find a policy that optimizes the behavior \cite{deisenroth13surveypolicysearch}. The balance between the amount of exploration necessary to learn the model and the exploitation of the latter to collect reward is, in general, an intractable problem and is usually called the exploitation-exploration dilemma.

Partial-observable markov decision processes (POMDP) generalize the concept for cases where the state is not directly observable \cite{kaelbling1998planning}.

\subsection{Space of Exploration Policies}
The policy space $\Pi$ is defined by all possible sequences of actions that can be taken by the agent or, alternatively, by all the different closed-loop policies that generate such sequences. 
The simplest approach is to select a single data point from the environment database and use it to improve the model $\hat{g}()$. In this case, $\Pi$ is defined by the set of all possible sequences of data points (or the algorithm, or sensor, that is used to select them).
Another case is when autonomous agents gather information by moving in the environment. Here, the actions usually include all the trajectories necessary to sample particular locations (or the motion commands that take the agent to them). 

\begin{figure}[htbp]
	\centering
		\includegraphics[width=0.40\columnwidth]{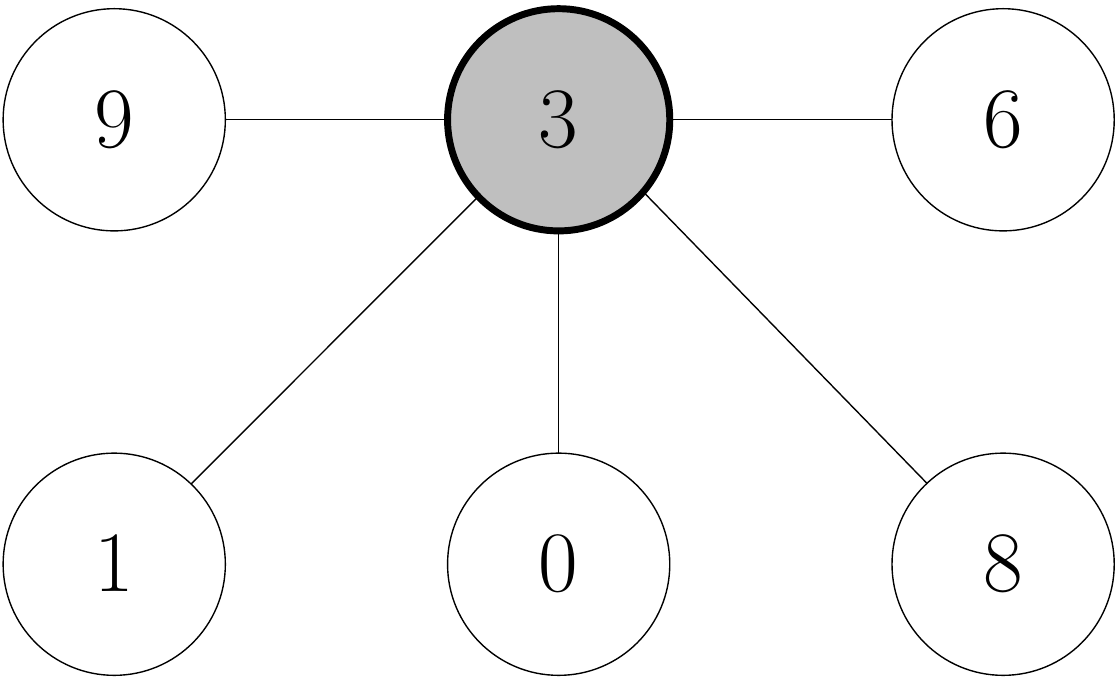}\hfill
		\includegraphics[width=0.40\columnwidth]{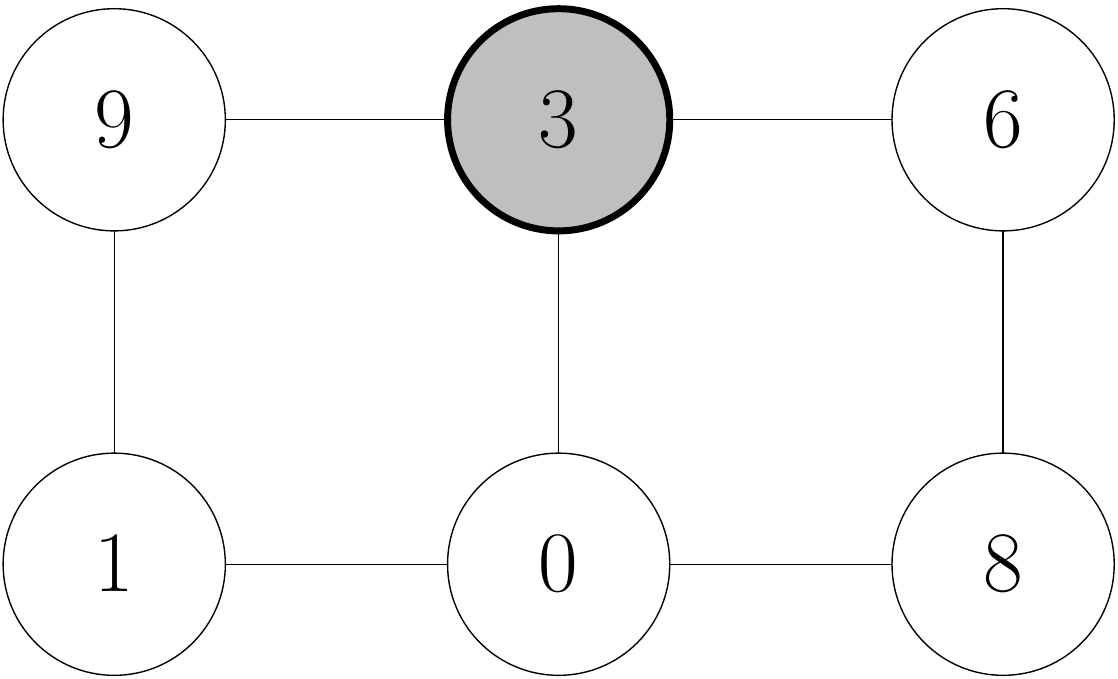}
	\caption{Different possible choices available to an exploring agent. Considering an agent in an initial state (grey state) it has to decide where to explore next (information value indicated by the values in the nodes). From the current location all the states might be reachable  (Left figure), or there might be restrictions and some state might might only be reachable after several steps (Right figure). In the latter case the agent has also to evaluate what are the possible actions after each move.}
	\label{fig:pathplanchoices}
\end{figure}

However, the formulation of Eq. \ref{eq:costopt} is much more general and can incorporate any other possible decision to be made by the agent. An agent might try to select particular locations to maximize information or could select at a more abstract level between different regions, e.g. starting to map the beach or the forest. This idea can be pushed further. The agent might decide among different exploration types and  request a helicopter survey of a particular location instead of measuring with its own sensors. In this case the robot selects among different exploration types. 
The agent might even decide between learning methods and representations that, in view of the current data, will behave better, produce more accurate models or result in better performance (see Section \ref{sec:selfguidedexplor}). This choice modifies the function $\hat{g}()$ used to compute the loss and can be changed several times during the learning process. 

The following list summarizes the possible choices that have been considered in the literature in the context of active learning:

\begin{itemize}
	\item next location, or next locations
	\item among a pre-defined partition of the space
	\item among different exploration algorithms
	\item learning methods
	\item representations
	\item others
\end{itemize}

\subsection{Cost}
\label{sec:Cost}

The term $C(\pi)$ represents the cost of the policy and we will assume that each action $a_t$ taken following $\pi$ incurs a cost which is independent of future actions. However, the cost of an action may depend on the history of actions and states of the agent. Indeed, modeling this dependency is an important design decision, specially for autonomous agents. Figure \ref{fig:pathplanchoices} illustrates the implications of this dependency. In the first example, the cost of an action $C(a_t)$ depends only on the action. This is usually the case of costs associated to sensing the environment.  In the second case, the cost $C(a_t \mid a_{t-1})$ depends on the previous action since it implies a non-zero cost motion. This type of cost appears naturally for autonomous agents that need to move from one location to another\footnote{Action $a_t$ is not precisely defined yet. The previous distinction abuses notation by abstracting over the specific action definition (e.g. local displacements or global coordinates). The important thing is that moving incurs a cost that depends on previous actions.}. In many cases, the cost will consist of a combination of different costs that can individually depend or not on previous actions.

\subsection{Loss and Active Learning Tasks}

\begin{table*}
	\centering
	\caption{Taxonomy active learning}
		{\small
		\begin{tabular}{|c|c|c|}\hline
Choice$\backslash$ Prob.  	&	Optimization 							& Learning \\\hline
Point			&	Bayesian Optimization \cite{brochu2010tutorial}							& Classical Active Learning \cite{Settlesactivelearsurv}	\\\hline
Discrete tasks	& Multi-armed bandits \cite{auer2003nonstochastic}																	& AL for MAB	\cite{Carpentier11ActMAB}		\\\hline
Trajectory			& Exploration/Exploitation \cite{Kaelbling96jmlr}				& Exploration \\\hline			
		\end{tabular}}
	\label{tab:TaxonomyActiveLearning}
\end{table*}

The term $\LL_{x\in \mathcal{X}}(\hat{g}(x; D \sim \pi))$ represents the loss incurred by the exploration policy. Recall that the agent's objective is to learn model $g()$. The loss is therefore defined as a function of the goodness of the learned model. Obviously, the function $g()$ varies with the task. It can be a discriminant function for classification, a function approximation for some quantity of interest or a policy mapping states to actions. In any case, the learned function $\hat{g}()$ will be determined by the flow of observations $D$ induced by the policy $\pi$ (e.g. training examples for a classifier or measurements of the environment to build a map). 


Another important aspect that must be considered is when the loss is evaluated. One possibility is that only the final learned model $\hat{g}()$ is used to obtain the expected loss. In this case, mistakes made during {\em training} are not taken into account. Alternatively, one may consider the accumulated loss during the whole lifetime of the agent, where even the cost and errors made during the learning phase are taken into account. We can also think that no explicit learning phase exists in this setting. In the MAB literature these measures are known as the \textit{simple regret} and \textit{average regret}. The latter tells, in hindsight, how much was lost by not pulling always the best arm. And the former tells how good is the arm estimated as being the best.

Earlier on, we did not make explicit what the loss function aims to capture during the learning process. 
Again, there are two possible generic options to consider: learn the whole environment (what we consider to be a pure \textit{learning} problem); or find a location of the environment that provides the highest value (\textit{optimization} problem). 
Note that in both cases, it is necessary to learn a model $g()$. However, in the first case we are interested in minimizing the error of the learned model
\begin{equation}
\int_{\mathcal{X}} \LL(g(x),\hat{g}(x))dx
\label{eq:purelearning}
\end{equation}
while in the second case we are just interested on fitting a function $g()$ that helps us to find the maximum of
\begin{equation}
max_x g(x)-g\left(argmax_x \hat{g}(x)\right)
\label{eq:probmax}
\end{equation}
irrespectively of what the function $\hat{g}()$ is actually approximating. 
In a multi-armed bandit setting this amounts to just detect which is the best arm, or learn the payoff of all the arms. Table \ref{tab:TaxonomyActiveLearning} summarizes this perspective. In this \textit{pure learning} problem of multi-armed bandits regret bounds on the simple regret can also be made \cite{Carpentier11ActMAB,gabillon2011MABident}. For the general RL problem regret bounds have also been established \cite{jaksch2010near}.

\subsection{Measures of Information}
\label{sec:Information}

The utility of the policy in Eq. \ref{eq:costopt} is measured using a function $f()$. Computing the information gain of a given sample is a difficult task which can be computationally very expensive or intractable.
Furthermore, it can be implemented in multiple different ways depending on how the information is defined and on the assumptions and decisions done in terms of loss, cost and representation. 
Also, we note that in some cases, due to interdependencies between all the points, the order in which the samples are obtained might be relevant.
The classification below follows the one proposed in  \cite{Settlesactivelearsurv} (also refer to \cite{mackay1992information,Settlesactivelearsurv} for further details) and completes it by including empirical measures as a different way of assessing the information gain of a sample. The latter class of measures aims to consider those cases where there is no single model that covers the whole state-space, or if the agents lacks the knowledge to select which is the best one \cite{Schmidhuber91curiosity,oudeyer2007intrinsic}.

\subsubsection{Uncertainty sampling and Entropy}
The simplest way to select the new sample is to select the one we are currently more uncertain about. Formally, this can be modeled as the entropy of the output. Uncertainty sampling where the query is made where the classifier is most uncertain about \cite{lewis1994sequential}, still used in support vector machines \cite{Tong01}, logistic regression \cite{Schein07}, among others.

\subsubsection{Minimizing the version space}
\label{sec:VersionSpaceModels}

The version space defines the subset of all possible models (or parameters of a model) that are consistent with the current samples and, therefore, provides the set of hypotheses we are still undecided about. This space cannot in general be computed.  It has been approximated in many different ways. An initial model considered \textit{Selective Sampling} \cite{cohn1994improving} where a pool, or stream, of unlabeled examples exists and the learner may request the labels to an oracle. The goal was to minimize the amount of labeled data to learn the concepts to a fixed accuracy. \textit{Query by committee} \cite{seung1992query,freund1997selective} considers a committee of classifiers and measures the degree of disagreement between the committee. 
Another perspective was proposed by \cite{Angluin88queries} to find the correct hypothesis using \textit{membership queries}. In this method the learner as a class of hypothesis and has to identify the correct hypothesis exactly. Perhaps the best-studied approach of this kind is learning by queries \cite{Angluin88queries,cohn1994improving,baum1991neural}.  Under this setting approaches have generalized methods based on binary search \cite{nowak2011geometry,Melo13GBS}.
Also, active learning in support vector machines can be seen in a version space perspective or as the uncertainty of the classifier \cite{Tong01}.

\subsubsection{Variance reduction}
\label{sec:StatisticalModels}

Variance reduction aims to select the sample(s) that will minimize the variance of the estimation for unlabeled samples \cite{Cohn96}.  There exist closed form solutions for some specific regression problems (e.g. linear regression or Gaussian mixture models). In other cases, the variance is computed over a set of possible unlabeled examples which may be computationally expensive. 
Finally, there are other decision-theoretic based measures such as the expected model change \cite{settles2007multiple} or the expected error reduction \cite{roy2001toward,moskovitch2007improving} which select the sample that, in expectation, will result in the largest change in the model parameters or in the largest reduction in the generalization error, respectively.

\subsubsection{Empirical Measures}
\label{sec:EmpiricalMeasures}  

Empirical measures make less assumptions on the data-generating process and instead estimate empirically the expected quality of each data-points/region \cite{Schmidhuber91curiosity,Schmidhuber06,oudeyer2007intrinsic,Oudeyer2007,Lopes12zeta}. This type of measures consider problems where (parts of-) the state space have properties that change over time, can not be learned accurately, or require much more data than other parts given a particular learning algorithm. 
Efficient learning in those situations will require to balance exploration so that resources are assigned according to the difficulty of the task. 
In those cases where this prior information is available, it can be directly incorporated in the previous methods. The increase in complexity may result in computationally expensive algorithms. When the underlying structure is completely unknown, it might be difficult to find proper models to take into account all the uncertainty. And even for the case where there is a generative model that explains the data, its complexity will be very high. 

Let us use a simple analogy to illustrate the main idea behind empirical measures. Signal theory tells us what is the sampling rate required to accurately reconstruct a signal with a limited bandwidth. To estimate several signals, an optimal allocation of sampling resources would require the knowledge of each signal bandwidth. Without this knowledge, it is necessary to estimate simultaneously the signal and the optimal sampling rate, see Figure \ref{fig:actmabprogress}. Although for this simple case one can imagine how to create such an algorithm, the formalization of more complex problems might be difficult. 
Indeed, in real applications it is quite common to encounter similar problems. For instance,  a robot might be able to recover the map in most parts of the environment but fail in the presence of mirrors. Or, a visual attention system might end up spending most of its time looking at a tv set showing static. 

The first attempt to develop empirical measures was made by \cite{schmidhuber1991curious,Schmidhuber91curiosity} in which an agent could model its own expectation about how future experiences can improve model learning. After this seminal paper, several measures to empirically estimate how can data improve task learning have been proposed and a integrated view can be seen in \cite{Oudeyer2007}. To note that if there is an accurate generative model of the data, then empirical measures reduce to standard methods, see for instance the generalization of Rmax method \cite{brafman2003rmax} to the use of empirical measures in \cite{Lopes12zeta}. 

\begin{figure}[htbp]
	\centering
		\includegraphics[width=0.8\columnwidth]{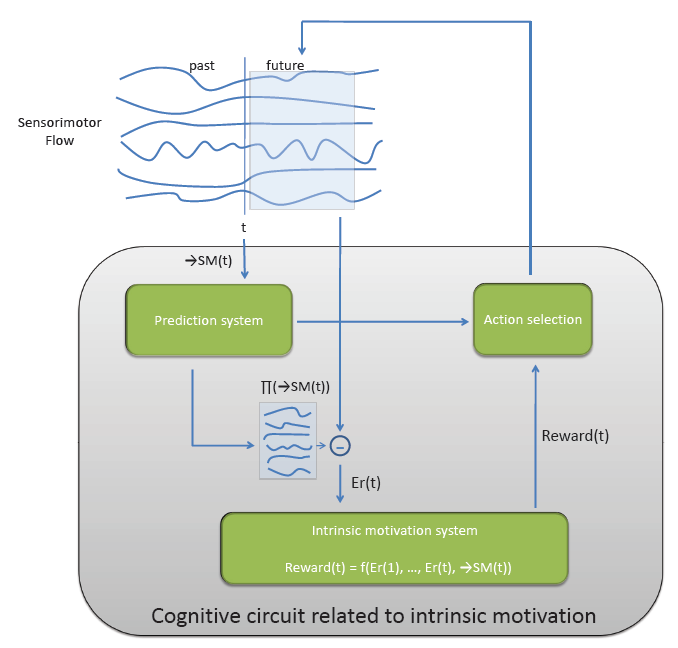}
	\caption{Intrinsic motivation systems rely on the use of empirical measure of learning progress to select actions to promise higher learning gains. Instead of considering complex statistical generative models of the data, the actual results obtained by the learning system are tracked and used to create an estimator of the learning progress. From \cite{Oudeyer2007}.}
	\label{fig:intrinsmot}
\end{figure}

In more concrete terms empirical measure rely not on the statistical properties of a generative data model, but on tracking the evolution of the quality of estimation, see Figure \ref{fig:intrinsmot}.

A simple empirical measure of learning progress $\zeta$ can be made by estimating the variation of the estimated prediction error. If we consider a loss model $\LL$ for the learning problem as: $\LL(\hat \TT; D)$, where $\TT$ is the true model and $D$ is the observed data. Putting an absolute threshold directly on the loss is hard. Note that the predictive error has the entropy of the true distribution as a lower bound, which is unknown \cite{Cohn96}. Therefore, these methods drive exploration based on the learning
\emph{progress} instead of the current learner accuracy. Using the
change in loss they may gain robustness by becoming independent of the
loss' absolute value and can potentially detect time-varying
conditions \cite{Oudeyer2007,Lopes12zeta}.

We can define $\zeta$ in terms of the change in the (empirically estimated)
loss as follows. Let $D^{-k}$ denote the experiences in $D$ except the last
$k$ and $\hat \TT^{-k}$ is the transition model learned from the reduced
data-set $D^{-k}_{s,a}$. We define
$\hat\zeta \approx \LL(\hat \TT^{-k}; D) - \LL(\hat \TT; D)$.
This estimates to which extent the last $k$ experiences help to
learn a better model as evaluated over the complete data. Thus, if
$\hat\zeta$ is small, then the last $k$ visitations in the
data-set $D$ did not have a significant effect on improving $\hat
\TT$. To note that finding a good estimator for the expected loss is not trivial and resampling methods might be required \cite{Lopes12zeta}. See also \cite{Oudeyer2007} for different definitions of learning progress.

\subsection{Solving strategies}
\label{sec:HowToExplore}

The optimal exploration problem defined in  Eq. \ref{eq:costopt} is in its most  general case computationally intractable. Note that we aim at finding a exploration policy, or an algorithm, that is able to minimize the amount of data required while minimizing the loss. In Fig. \ref{fig:3perspectives} that would amount to choose among all the possible trajectories, of equivalent cost, the ones that provide the best fit.
%
%
Furthermore, common statistical learning theory does not directly apply to most active learning algorithms and it is difficult to obtain theoretical guarantees about their properties. The main reason is that most theory on learning relies on the assumption that data is acquired randomly, i.e. the training data comes from the some distribution as the real data, while in active learning the agents itself chooses the next data point.

\subsubsection{Theoretical guarantees for binary search}

Despite previous remarks, there are several cases where it is possible to show that active learning provides a gain and obtain some guarantees. \cite{Castro2008,Balcan2008} identify the expected gains that active learning can give in different classes of problems. For instance, \cite{dasgupta05greedyact,dasgupta2011two} studied the problem of actively finding the optimal threshold on a line for a separable classification problem. A binary search applied to this problem yields an exponential gain in sample efficiency.  In what conditions, and for which problems this gain still hold is currently under study. As discussed by the authors, in the worst case it might still be necessary to classify the whole dataset to identify the best possible classifier. However, if we consider the average case and consider the expected learning quality for finite sample sizes, results show that we can get exponential improvements over random exploration. Indeed, other authors have shown that generalized binary search algorithms can be derived for more complex learning problems \cite{nowak2011geometry,Melo13GBS}.

\subsubsection{Greedy methods}

Many practical solutions are greedy, i.e. they only look at maximizing directly a function. We note the difference between a greedy approach that directly maximizes a function an a myopic approach that ignores the long-term effects of those choices. As we discuss now, there are cases where greedy methods are not myopic.
The question is how far are greedy solutions from the optimal exploration strategy. This is in general a complex combinatorial problem. If the loss function being minimized has some structural properties, then some guarantees can be found that relate the sample complexity of a given algorithm with the possible best polynomial time algorithm. 
Under this approach the submodular property has been extensively used \cite{krause2005near,golovin10near,golovin10adaptive,maillard2012hierarchical}. Submodular functions are functions that observe the diminishing return property, i.e. if $B \subset A$ then $F(A\cup \{x\}) - F(A) \geq F(B\cup \{x\})-F(B)$. This means that choosing a datapoint sooner during the optimization will always provide equal or more information than the same point later on. 

A theorem from \cite{nemhauser1978analysis} says that for monotonic submodular functions, the value of the function for the set obtained with the greedy algorithm $G(D_g)$ is close, $(1-1/e)$, to the value of the optimal set $G(D_{OPT}$). This means that if we would solve the combinatorial problem, the solution we get with the greedy algorithm is at most $33\%$ below the true optimal solution. 

Unfortunately not all problems are submodular. First, some target functions are not submodular. Second, online learning methods introduce bias since the order of the data changes the active learning results. Third, some problems cannot be solved using a greedy approach. For these problems a greedy algorithm can be exponentially bad (worst than random exploration). Also, a common situation is to have submodular problems given some unknown parameters without which it is not possible to use a the greedy algorithm. In this situation it is necessary to take an exploration/exploitation strategy to explore the parameter space to gather information about the properties of the loss function and and then exploit it.

\subsubsection{Approximate Exploration}

The most general case as shown in Figure \ref{fig:3perspectives} is not submodular and the best solution rely of PAC-bounds.  Two of the most influential works on the topics are: $E^{3}$ \cite{kearns2002near} and \rmax\ \cite{brafman2003rmax}. Both take into account how often a state-action pair has been visited to decide if further exploration is needed or if the model can be trusted enough (in a PAC setting) to be used for planning purposes. With different technical details both algorithms guaranty that with high-probability the system learns a policy whose value is close to the optimal one. Some other approaches consider limited look-ahead planning to approximately solve this problem \cite{sim05,Krause07}.

%

\subsubsection{No-regret}

In the domain of multi-armed bandits several algorithms have been developed that can solve the optimization \cite{gabillon2011MABident} or the learning \cite{Carpentier11ActMAB} problem with the best possible regret sometime taking into account specific knowledge about the statistical properties of each arm, but in many cases taken a distribution free approach \cite{auer2003nonstochastic}.

	\section{Exploration}
\label{sec:selfguidedexplor}

In this section we present the main approaches of active learning, particularly focused in systems with physical restrictions, i.e. where the cost depends on the state. This section organizes the literature according to what is being selected as policy for exploration. The distinctions are not clear in some cases, and some works include aspects of more than one problem, or can be seen in different perspectives. We consider three different parts: greedy selection of points where $C(a_t|a_{t-1})=C(a_t)$ and considering a selection among an infinite set of points or among a finite set, the last part considers the cases where the selection takes explicitly into account $C(a_t|a_{t-1})$ and longer time horizons. There is already a great variety of approaches but mainly the division corresponds to classical active learning, multi-armed bandits and exploration-exploitation in reinforcement learning. We are interested in applications related to general autonomous agents and will only consider approaches focused on the classical active learning methods if they provide a novel idea.

%
\subsection{Single-Point Exploration}
\label{sec:MyopicExploration}

\begin{figure}
	\centering
		\includegraphics[width=0.25\columnwidth]{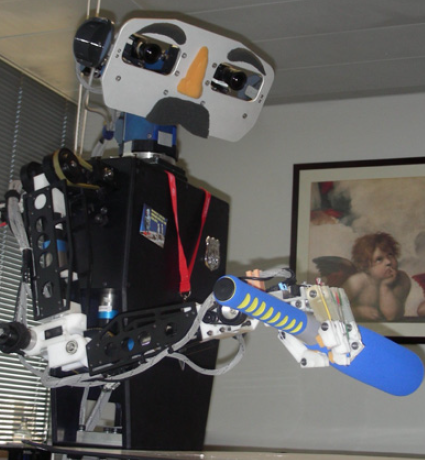}
		\includegraphics[width=0.73\columnwidth]{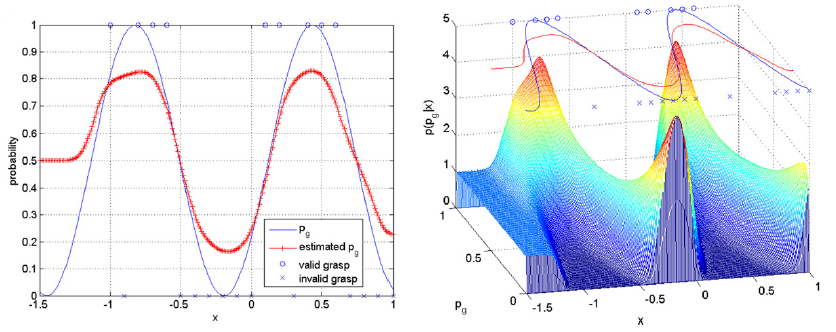}
	\caption{Approximating a sinus varying p in a one dimensional input space representing a robot actively learning which object locating afford a more successful grasp. (a) Robotic setup. (b) Estimated mean. The blue points are the observations generated from a Bernoulli
experiment, using the true p (blue line). Failures are represented by crosses and successes by circles. The red line with marks is the approximated mean computed from the
posterior. (b) Predicted posterior beta distributions for each point along x. From \cite{montesano12actbetas}.}
	\label{fig:beta}
\end{figure}

This section describes works that, at each time step, choose which is the single best observation point to explore without any explicit long term planning. This is the most common setting in active learning for function approximation problems \cite{Settlesactivelearsurv}, with examples ranging from vehicle detection \cite{sivaraman2010general}, object recognition \cite{Kapoor07} among others. Note that, as seen in Section \ref{sec:HowToExplore}, in some cases information measures were defined where a greedy choice is (quasi-) optimal. Figure \ref{fig:beta} provides an example of this setting where a robot is able to try to grasp an object at any point to learn the probability of success, at each new trial the robot can still choose amongst the same (infinite) set of grasping points.

\subsubsection{Learning reliability of actions}
\label{sec:LearningReliabilityOfActions}

An example of the use of active learning under this setting, and with particular interest for physical systems, is to learn the reliability of actions. For instance, it has been suggested that grasping could be addressed by learning a function that relates a set of visual features with the probability of grasp success when a robot tries to grasp at those points \cite{Ng06}. This process requires a large database of synthetically generated grasping points (as initially suggested by \cite{Ng06}), or alternatively to actively search and select where to apply grasping actions to estimate their success \cite{salganicoff96activelearninggrasp,Morales04}. Another approach, proposed by \cite{Montesano09,montesano12actbetas} (see also Figure \ref{fig:beta}), derived a kernel based algorithm to predict the probability of a successful grasp together with its uncertainty based on Beta priors. Another approach used Gaussian process to model directly probability densities of successful grasps \cite{Detry09}. Clearly such success probabilities depend on the grasping policy is being applied, and a combination of the two will be required to learn the best grasping strategy \cite{kroemer2009active,kroemer2010combining}.
 
Another example is to learn several terrain properties in mobile robots such as obstacle detection and terrain classification. \cite{dima2004enabling} use active learning to request human users the correct labels of extensive datasets acquired by robots using density measures. Also using multiview approaches \cite{dima2005active}. Another property exploited by other authors is the traversability of given regions \cite{ugur2007curiosity}.

A final example considers how to optimize the parameters of a controller whose results can only be evaluated as success or failure \cite{tesch13expensive}. The authors rely on Bayesian optimization to select which parameters are still expected to provide higher probabilities of success.

\subsubsection{Learning general input-output relations}
\label{sec:LearningGeneralInputOutputRelations}

Several works explore different ways to learn input-ouputs maps. A simple case is to learn forward-backward kinematic or dynamical models of robots but it can also be the effects of time extended policies such as walking.

To learn the dynamical model of a robot, \cite{Cantin10icra} considered how to select which measure to gather next to improve the model. The authors consider a model parameterized by the location and orientation of a rigid body and their goal is to learn such parameters as fast as possible. They rely on uncertainty measures such as a-optimality. 

For non-parametric models several works learn different models of the robotic kinematic, using either nearest-neighbors \cite{baranes2013active} or local-linear maps \cite{rolf2011online}. Empirical measures of learning progress were used by \cite{baranes2013active} and \cite{rolf2011online}.



\subsubsection{Policies}
\label{sec:Policies}
Another example is to learn what action to apply in any given situation. In many cases this is learned from user input. This setting will be discussed in detail in Section \ref{sec:LearningByDemonstration}.

\cite{Chernova09jair} considering support vector machines as the classification method. The authors consider the confidence on the prediction of the SVM and while the robot is moving it will query the teacher when that confidence is low.

Under the formalism of inverse reinforcement learning, queries are made to a user that allow to infer the correct reward \cite{macl09airl,melo2010metric,cohn2010selecting,cohn2011comparing,judah12active}. Initial sample complexity results show that this approaches can indeed provide gains on the average case \cite{Melo13GBS}.

\subsection{Multi-Armed Bandits}
\label{sec:MAB}

This section discusses works that, similarly to the previous section, solely choose a single exploration point. The main difference is that we consider here the setting where this choice is discrete, or categorical. There are several learning problems that fall under this setting: environmental sensing and online sensor selection, multi-task learning, online selection of learning/exploration strategy, among others (see Table \ref{tab:mabexplor}). 

\begin{figure}[htbp]
	\centering
		\includegraphics[width=0.33\columnwidth]{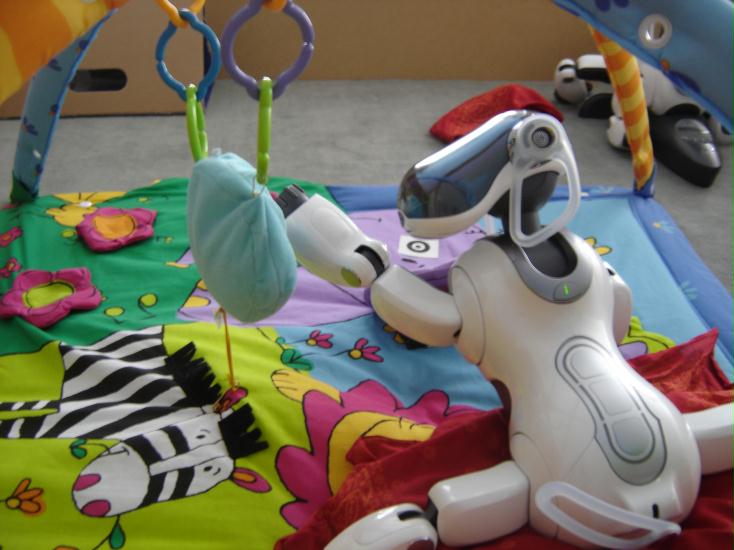}
		\includegraphics[width=0.65\columnwidth]{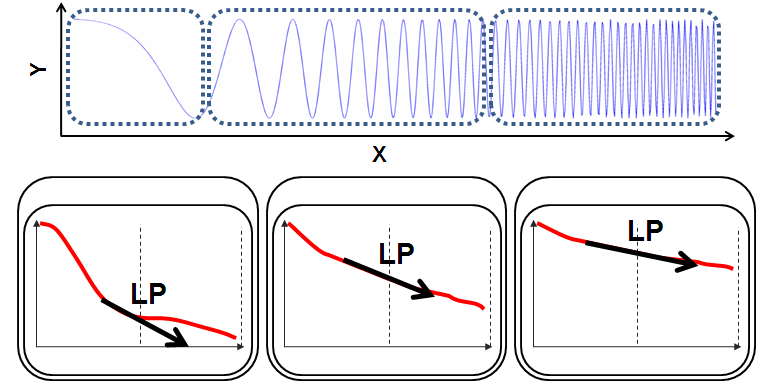}
	\caption{An example of a regression problem where the properties of the function to be learned vary along space. An optimal sampling of such signal will be non-uniform and could be solved efficiently if the signal properties were known. Without such information exploration strategies must be devised that learn simultaneously the properties of the signal and sample it efficiently. See \cite{Lopes12ssp} for a discussion. From \cite{oudeyer2007intrinsic}.}
	\label{fig:actmabprogress}
\end{figure}

There are two main origins for this different set of choices. One is that the problem is intrinsically discrete. For instance the system can either be able to select among a set of different sensors, different learning algorithms \cite{baram2004online,hoffman2011portfolio,Hester13aamas}, or being interested in learning from among a set of discrete tasks \cite{barto2004intrinsically}. Another case is when the discretization is made to simplify the exploration problem in a continuous space, reducing the cases presented in Section \ref{sec:MyopicExploration} to a MAB problem. Examples include  environmental sensing where the state is partitioned for computational purposes \cite{Krause08}, or learning dynamical models of robots where the partition is created online based on the similarities of the function properties at each location \cite{oudeyer2005playground,baranes2009riac} (see Figure \ref{fig:actmabprogress}). In all cases the goal is to learn a function in all domain by learning a function in each partial domain. Or to learn the relation of all the choices with their outputs. For a limited time horizon the best overall learning must be obtained.

In the recently introduced \textit{strategic student problem} \cite{Lopes12ssp}, the authors provide an unified view of these problems, following a computational approach similar to \cite{baram2004online,hoffman2011portfolio,baranes2013active}. After having a finite set of different possible choices that can be explored, both problems can be approached in the same way and relying on variants of the EXP4 algorithm \cite{auer2003nonstochastic}. This algorithm considers adversarial bandit settings and relies on a collection of experts. The algorithm has zero regret on the choice of experts and each expert will track the recent quality of each choice. 

We note that most algorithms for MAB were defined for the exploration-exploitation setting, but there are cases where there is a pure-exploration problem. The main difference is that if we define the learning improvement as reward, this reward will change with time, as sampling the same location will reduce its value. It is worth to note that if the reward function were known then most of these cases could be reduced to a submodular optimization where a greedy heuristic is quasi-optimal. When this is not the case then a MAB algorithm must be used to ensure proper exploration of all the arms \cite{Lopes12ssp,golovin10online}. 

One interesting aspect to note is that, in most of cases, the optimal strategy is non-stationary. That is, for different time instants, the percentage of time applied to each choice is different. We can see that there is a developmental progression from learning simpler topics to more complex ones. Even at the extreme cases where with little amount of time some choices are not studied at all. These results confirms the heuristics of learning progress given by \cite{Schmidhuber91curiosity,Oudeyer2007}. Both works considered that at any time instant the learner must sample the task that has given a larger benefit in the recent past. For the case at hand we can see that the solution is to probe, at any time instant, the task whose learning curve has an higher derivative, and for smooth learning curves both criteria are equivalent.

We will now present some works that do active exploration by selecting among a finite set of choices. We divide the approaches in terms of choosing different (sub-) tasks or different strategies to explore, or learn a single task. Clearly this division depends on different nomenclatures and on how the problems are formulated.

\subsubsection{Multiple (Sub-)Tasks}

In this case we considered that there is a set of possible choices to be made that correspond to learning a different (sub-)task. This set can be pre-defined, or acquired autonomously (see Section \ref{sec:Curiosity}), to have a large dictionary of skills that can be used in different situations or to create complex hierarchical controllers \cite{barto2004intrinsically,RByrneImitativeMind}

Multi-task problems have been considered in classification tasks \cite{qi2008two,reichart2008multi}. Here active learning methods are used to improve not only one task, but the overall quality of the different tasks.

More interestingly for our discussion are the works from \cite{Singh2005intrinsically,Oudeyer2007}. The authors divide the problem of learning complex agent-environment tasks into learning a set of macro-action, or predictive models, in an autonomous way (see Section \ref{sec:Curiosity}). These initial problems took very naive approaches and were latter improved with more efficient methods. \cite{Oudeyer2007} initially considered that each parameter region gave a different learning gain, and the one that were given the highest gain was selected. Taking into account the previous discussion we know that a better exploration strategy must be applied and the authors considered more robust measures and created a stochastic policy to provide efficient results in high-dimensional problems \cite{baranes2013active}. More recently \cite{maillard2012hierarchical} introduce a new formulation of the problem and a new algorithm with specific regret bounds. The initial work of \cite{Singh2005intrinsically} lead to further improvements. The measures of progress that guide the selection of the macro action that is to be chosen started to consider the change in value function during learning \cite{csimcsek2006intrinsic}. Similar ideas were applied to learn affordances \cite{hart2013intrinsically} where different controllers and their validity regions are learned following their learning progress.

In distributed sensing it is required to estimate which sensors provide the most information about a environmental quantity. Typically this quantity is time varying and the goal is to actively estimate which sensors provide more information. When using a gaussian process as function approximation it is important to consider exploration to find the property of the kernel and then, for known parameters of the kernel, a simple offline policy provides optimal results \cite{Krause07}. This partition in a finite set of choices allows to derive more efficient exploration/sensing strategies and still ensure tight bounds \cite{Krause08,golovin10adaptive,golovin10online}.

\subsubsection{Multiple Strategies}

The other big perspective is to consider that the choices are the different methods that can be used to learn from the task, in this case a single-task is often considered. This \textit{learning how to learn} approach makes explicit that a learning problem is extremely depending on the method to collect the data and the algorithm used to learn the task. 

Other approaches include the choice among the different teachers that are available to be observed \cite{price03rlimitation} where some of them might not even be cooperative \cite{Shon07imitmultdemons}, or even choose between looking/asking for a teacher demonstration or doing self-exploration \cite{mai2012choice}.

Another approach considers the problem of having different representation and selecting the best one. The representation that gives more progress will be used more frequently \cite{konidaris2008sensorimotor,Maillard11SelState}.

The previous mentioned work of \cite{Lopes12ssp} showed also that the same algorithm can be used to select online which exploration strategy was best to learn faster the transition probability model of an MDP. The authors compared R-Max, $\epsilon-greedy$ and random. A similar approach was suggested by \cite{castronovo2012learning} where a list of possible exploration reward is proposed and a given arm bandit is assigned to each one. Both works took a simplified approach by considering that reset actions were available and the choices were only made at the beginning of each episode. This limitation was recently improved by considering that the agent can evaluate and select online the best exploration strategies \cite{Hester13aamas}. In this work the authors relied on a factored representation of an MDP \cite{hester12modellearnintrin} and using many different exploration bonuses they were able to define a large set of exploration strategies. The new algorithm at each instant computes the gain in reward for the selected exploration strategy and simultaneously the expected gain for all the other strategies using an importance sampling idea. Using such expected gains the system can select online the best strategy given better results than any single exploration strategy would do.
 
\begin{table*}
	\centering
	\caption{Formulation of several Machine Learning problems as a Strategic Student Problem.}
	{\small
		\begin{tabular}{lllll}
			Prob. &Choices		&Topics		&References\\\hline			
			reg. 		& n Regions														&n Functions						&	\cite{baranes2010goalexplor,baranes2013active}\\
			mdp 		& n Environment											&n Environments			&	\cite{barto2004intrinsically,oudeyer2005playground,Oudeyer2007}\\
			reg. 		& n Environment											&n Environments			&	\cite{Lopes12ssp}\\
			reg.		& Control or Task Space						&Direct/Inv. Model & \cite{baranes2013active,Jamone11explor,rolf2011online}\\
			\hline
			mdp 		& Exploration strategies						&1 Environment				&	\cite{baram2004online,Krause08,Lopes12ssp}\\
			mdp 		& n Teachers													&1 Environment				&	\cite{price03rlimitation,Shon07imitmultdemons}\\
			reg. 		& Teacher,self-exploration 	&1 Function							&	\cite{mai2012choice}\\
			mdp 		& n Representations 								&1 Environment				&	\cite{konidaris2008sensorimotor,Maillard11SelState}\\
		\end{tabular}	}
	\label{tab:mabexplor}
\end{table*}

\subsection{Long-term exploration}
\label{sec:LongTermExploration}

\begin{figure}[htbp]
	\centering
		\includegraphics[width=1.00\columnwidth]{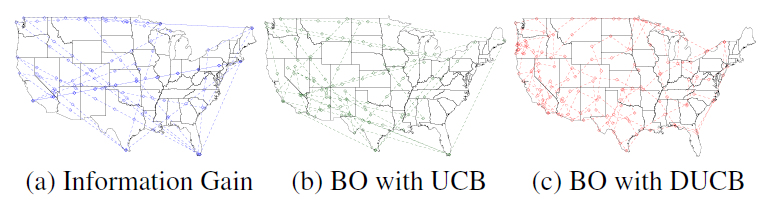}
	\caption{In environmental monitoring it is necessary to find the trajectories that provide the more critical information about different variables. Selecting the most informative trajectories based on the space and time variation and the physical restrictions on of the mobile sensors is a very complex problem. The figures show the trajectories followed by simulated aerial vehicles, samples are only allowed inside the US territory. Courtesy from \cite{marchant2012bayesian}.}
	\label{fig:figactplanenvmon}
\end{figure}

We now consider active exploration strategies in which the whole trajectory is considered within the optimization criteria instead of planning only a single step ahead. A real world example is the one of selecting informative paths for environmental monitoring, see Figure \ref{fig:figactplanenvmon}.

We divide this section in two parts. A first part entitled \textit{Exploration in Dynamical Systems} considering exploration where the dynamical constraints of the system are taken into account and another, that considers similar aspects, specific to \textit{Reinforcement Learning} and \textit{Markov Decision Processes}. We make this distinction due to the different communities, formalisms and metrics commonly used in each domain.

\subsubsection{Exploration in Dynamical Systems}
\label{sec:ExplorationInDynamicalSystems}

The most representative example of such a problem is one of the best studied problems in robotics: simultaneous localization and mapping (SLAM). The goal is to build a map of an unknown environment while keeping track of the robot position within it. 
Early works focused on active localization given an a priori map. In this case, the objective is to actively move the robot to obtain a better localization. In \cite{fox1998active} the belief over the robot position and orientation was obtained using a Monte Carlo algorithm. Actions are chosen based on a utility function based on the expected entropy of the robot location. A set of predefined relative motions are considered and only moving costs are considered. 

 
The first attempts to actively explore the environment during SLAM aimed to maximize the expected information gain \cite{feder99,Bourgault2002,Stachniss2003,Stachniss2005}. The implementation details depend on the on-board sensors (e.g. sonar or laser), the SLAM representation (feature based or grid maps) and the technique (EKF, Monte Carlo). For instance, in \cite{feder99} an EKF was used to represent the robot location and the map features measured using sonar. Actions were selected to minimize the total area of error ellipses for the robot and each landmark, by reducing the expected covariance matrix at the next time step.
For grid maps, similar ideas have been developed using mutual information \cite{Stachniss2003} and it is even possible to combine both representations \cite{Bourgault2002} using a weighted criteria. 
Most of the previous approaches consider just a single step ahead, have to discretize the action space or ignore the information that will be obtained during the path and its effect in the quality of the map.
A more elaborated strategy was proposed in \cite{sim05} where an a-optimality criterion over the whole trajectory was used. To make the problem computationally tractable, only a set of predefined trajectories is considered using breadth-first search. 
The work in \cite{ruben07actipollearn} directly aims to estimate the trajectory (i.e. a policy) in a continuous action-state space  taking into account the cost to go there and all the information gathered in the path \cite{ruben07actipollearn}. The policies are parameterized using way-points and the optimization is done over the latter. 
Some works explore similar ideas in the context of navigation and obstacle avoidance. For instance, \cite{kneebone2009navigation} uses a POMDP framework to incorporate uncertainty into Rapid Random Trees planning. The resulting policy takes into account the information the robot will obtain while executing the plan. Hence, the map is implicitly refined during the plan resulting in an improved model of the environment.

The active mapping approaches described above deal mainly with mapping environments with obstacles. However, similar ideas have been used to map other phenomena such as rough terrain, gas concentration or other environmental monitoring tasks. In this setting, robots allow to cover larger areas and to reconfigure the sensor network dynamically during operation. This makes active strategies even more relevant than in traditional mapping. Robots must decide where, when and what to sample to accurately monitor the quantities of interest.
In this domain it is important to consider learning non-stationary space-time models \cite{Krause07,garg2012efficient}. By exploiting submodularity it is possible to compute efficient paths for multiple robots assuring that they will gather information in a set of regions \cite{Singh07IJCAI}. Without relying on a particular division into regions, but without any proven bounds, \cite{marchant2012bayesian} used Bayesian optimization tools to find an informative path in a space-time model.

\subsubsection{Exploration / Exploitation}
\label{sec:ExplorationinRL}

Another setting where the learner actively plans its actions to improve learning is in reinforcement learning (see an early review on the topic \cite{thrun1992efficient}). In this general setting the agent is not just learning but is simultaneously being evaluated on its actions. This means that the errors made during learning count towards the global evaluation. In the Reinforcement learning (RL) approaches this is the most common setting. Under our taxonomy here the problem is also the one more challenging as the choice of where to explore next depends on the current location and the system has to take into account the way to travel to such locations.

As discussed before, this most general case, as shown in Figure \ref{fig:3perspectives}, is not submodular and there is not hope to find a computationally efficient method to solve it exactly. 
Initial proposals considered the uncertainty in the models and guided exploration based on this uncertainty and other measures such as recency of visits. The authors then proposed that a never-ending exploration strategy could be made that incorporates knowledge about already well know states and novel ones. \cite{schmidhuber1997reinforcement,Wiering98efficientmodelRL}. 

The best solutions, with theoretical guarantees, aim at finding efficient algorithms that have an high-probability of finding a solution that is approximately correct, following the standard probably approximately correct learning (PAC) \cite{strehl-littman:08-jcss,strehl-li-littman:09-jmlr}. Two of the most influential works on the topic are: $E^{3}$ \cite{kearns2002near} and \rmax\ \cite{brafman2003rmax}. Both take into account how often a state-action pair has been visited to decide how much further exploration must be done. Specifically, for the case of \rmax\ \cite{brafman2003rmax}, the algorithm divides the states into known and unknown based on the number of visits made. This number is defined based on general bounds for having a high certainty on the correct transition and reward model. Then the algorithm proceeds by considering a surrogate reward function that is \rmax\ in unknown states and the observed reward in known states. For a further analysis an more recent algorithm see the discussion in \cite{strehl-littman:08-jcss}.

PAC-RL measures consider that most of the times the agent will be executing a policy that is close to the optimal one. An alternative view is to see if the cumulative reward is close to the best one, as in the notion of regret. Such regret measure have been already generate some RL algorithms \cite{salganicoff1995active,ortner2007logarithmic,jaksch2010near}.

Yet another approach considers Bayesian RL \cite{dearden1998bayesianqL,poupart2006analbayesianRL,vlassis2012bayesian,sorg2010variance}. In this formalism the agents aims at finding a policy that is (close to) optimal taking into account the model uncertainty. The resulting policies solve implicitly the exploration-exploitation problem. Bayesian RL exploits prior knowledge about the transition dynamics to reason explicitly about the uncertainty of the estimated model. Bayesian exploration bonus (BEB) approach \cite{kolter2009beb} mixes the ideas of Bayesian RL with \rmax\ where the state are not explicitly separated between known and unknown but instead each state get a \textit{bonus} proportionally to the uncertainty in the model. The authors were able to show that this algorithm approximates the - hard to compute - bayesian optimal solution.

A recent approach considered how can \rmax\ be generalized for the case where each different state might have different statistical properties \cite{Lopes12zeta}. Specially in the case where the different properties are not known, empirical measures of learning progress have been proposed to allow the system to balance online the exploration necessary to verify the PAC-MDP conditions.

As a generalization of exploration methods in reinforcement learning, such as \cite{brafman2003rmax}, ideas have been suggested such as planning to be surprised \cite{sun2011planning} or the combination of empirical learning progress with visit counts \cite{hester12modellearnintrin}. This aspect will be further explored in Section~\ref{sec:Curiosity}.

We note also that the ideas and algorithms for exploration/exploitation are not limited to finite state representations, there have been recent results extending them to to POMDPs \cite{fox2007reinforcement,Jaulmes05actpomdp,Doshi08actpomdp}, Gaussian Process Dynamical Systems \cite{jung2010gaussian}, structured domains \cite{hester12modellearnintrin,nouri2010dimension}, and relational problems \cite{lang2010exploration}.

Most of the previous approaches are optimistic in the face of uncertainty. In the real world most of the times exploration must be done in incremental and safe ways due to the physical limits and security issues. In most cases process are not ergodic and care must be made. Safe exploration techniques have started to be developed \cite{Moldovan12safeexploration}. In this work the system is able to know if an exploration step can be reversed. This means that the robot can see ahead and estimate if it can return to the previous location. Results show that the exploration trajectory followed is different from other methods but allows the system to explore only the safe parts of the environment.


\subsection{Others}
\label{sec:Others}

There are other exploration methods that do not fit well in the previously defined structure, in most cases because they do not model explicitly the uncertainty. Relevant examples consider policy search and active vision. Other cases combine different methods  to accomplish different goals.

\subsubsection{Mixed Approaches}
\label{sec:MixedApproaches}

\begin{figure}
	\centering
		\includegraphics[width=1.00\columnwidth]{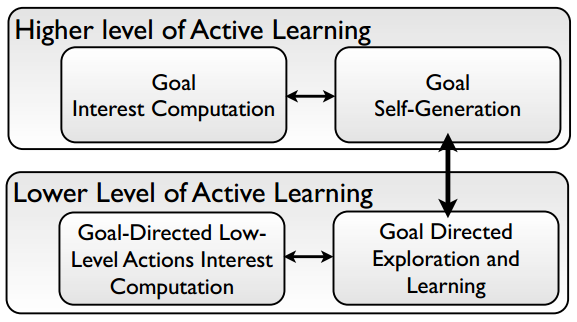}
	\caption{Several problem require the use of active learning at several different levels and/or time scales. Here is the examples of the SAGG-RIAC architecture. The structure is composed of two parts: a higher level for selecting target goals, and a lower level,
which considers the active choice of the controllers to reach such goals. The system allows to explore the space of reachable goals and learn the controllers required to reach them in a more efficient way. From \cite{baranes2013active}.}
	\label{fig:saggriac}
\end{figure}

There are several methods that include several levels of active learning to accomplish complex tasks, see Figure \ref{fig:saggriac}.

In \cite{MartinezCantin09AR,Cantin10icra} the authors want to learn a dynamical model of a robot arm, or a good map of the environment, with the minimum amount of data. For this it is necessary to find a trajectory, consisting of a sequence of via-points, that reduces the uncertainty on the estimator as fast as possible. The main difficulty is that this is in itself a computationally expensive problem, and if it is to be used in real time, then efficient Bayesian optimization techniques must be used \cite{brochu2010tutorial}.

Another examples is  the SAGG-RIAC architecture \cite{baranes2013active}. In this system a hierarchy of forward models are learned and for this it actively makes choices at two levels: in a goal space, it chooses what topic/region to sample (i.e. which goal to set), and in a control space, it chooses which motor commands to sample to improve its know-how towards goals chosen at the higher level.

We can also view the works of \cite{kroemer2009active,kroemer2010combining} as having a level of active exploration of good grasping points and another level of implicit exploration to find the best grasping strategies.

\subsubsection{Implicit exploration}
\label{sec:ImplicitExploration}

Learning in robots and data collection are always intertwined. Even if such data collection process is explicit in many cases, other situations, even if strongly dependent on that same process, address it only in an implicit way or as a side-effect of an optimization process \cite{deisenroth13surveypolicysearch}. The most noteworthy example are all policy gradient methods and similar approaches \cite{Sutton00nips,kober2013reinforcement}. In these methods the learner tries to directly optimize a policy given experiments and the corresponding associated reward. Some methods consider stochastic policies and the noise on the policy is used to perform exploration and collect data \cite{Peters05ecml}. The exploration reduces under the same process that adjust the parameters to improve the expected reward.
Another line of research is to use more classical methods of optimization to find the best set of parameters that maximize a reward function \cite{stulp2012policy}. Recently, and using a more accurate model of uncertainty it is possible to use Bayesian optimization methods to search for the best policy parameters that result in the highest success rate \cite{tesch13expensive}.

\subsubsection{Active Perception}
\label{sec:ActivePerception}

Another common use of the word active is in \textit{active perception}. Initially it was introduced because many computer vision problems become easier if more than one images is available or even a stream of video. An active motion of the camera can make such extra information much easier to discover. More recently it was motivated by the possibilities opened by having a robot acting in the environment to discover world properties.

This idea has been applied to segment object and learn about their properties \cite{metta03poking}, disambiguate and model articulated objects \cite{brock08bodyschema}, disambiguate sound \cite{berglund2005sound}, among others. Attention can also be seen as an instance of active perception, \cite{meger2008curious} presents an attention system and learning in a real environment to learn about object using SIFTs and finally, in highly cluttered environments active approach can also provide significant gains \cite{van2012maximally}.

\subsection{Open Challenges}
\label{sec:futureExploration}

Under the label of exploration we considered several domains that include standard active learning, exploration and exploitation problems, multi-armed bandits and general online learning problems. All these problems have already a large research body but there are still many open challenges.

Clearly a great deal of work is still necessary to expand the classes of problem that can be actively sampled in an efficient way.  In all the settings we described there exist already many different approaches, many of them with formal guarantees. Nevertheless, for any particular instance of a problem it is not clear what method is the most efficient in practice, or how to synthesize the exploration strategies from a problem domain description.

Some of the heuristics and methods, and also many of the hypothesis and models, proposed in the developmental communities can be natural extensions to the active learning setting. For instance there is a very limited research on active learning for more complex models such as time-variant problems, domains with heteroscedastic noise and properties (see many of the differences in Table \ref{tab:ActiveLearningVsIM}).

\section{Curiosity}
\label{sec:Curiosity}

Most active approaches for learning address the problem of learning a single, well defined, task as fast as possible. Some of the examples given, such as safe exploration, already showed that in many cases there is a multi-criteria goal to be fulfilled. In a truly autonomous and intelligent system knowing what tasks are worth exploring or even which tasks are to be learned is a ill-defined problem.

In the 50s and 60s researchers started to be amazed by the amount of time children and primates spend in tasks that do not have a clear objective return. This spontaneous motivation to explore and intrinsic curiosity to novelty \cite{berlyne1960conflict} challenged utilitarian perspectives on behavior. The main question is why do so many animals have a long period of playing and are curious, activities that in many perspectives can be considered risky and useless? One important reason seems to be that is this intrinsic motivation that will create situations for learning that will be useful in future situations \cite{baldassarre2011intrinsic,singh2009rewards}, only after going through school will that knowledge have some practical benefit. Intelligent agents are not myopically optimizing their behavior but also gathering a large set of perceptual, motor, and cognitive skills that will have a benefit in a large set of possible future tasks. A major problem is how to define a criteria of what a successful learning is if the task is just to explore for the sake of pure exploration. Some hypothesis can be made that this stage results from an evolutionary process that leads to a better performance in a class of problems \cite{Singh2010}.  Or that intrinsic motivation is a way to deal with bounded agents where maximizing the objective reward would be too difficult \cite{singh2010separating,sorg2010internal}. Even for very limited time spans where an agent wants to select a single action, there are many somewhat contradictory mechanisms for attention and curiosity \cite{gottlieb2012attention}. An agent might have preferences for: specific stimuli; actions to promise bigger learning gains; selecting actions that provide the required information for reward prediction/gathering.


The idea of assuming that the future will bring new unknown tasks can be operationalized even in a single domain.  Consider a dynamical environment (defined as a MDP) where there is a training phase of unknown length. In one approach the agent progressively learns how to reach all the states that can be reached in 1 step. After being sufficiently sure that it found all such states and has a good enough policy to reach them the system increases the number of steps and starts the process. This work, suggested by \cite{Auer11intrRL,lim2012autonomous}, shows that it is possible to address such problem and still ensure formal regret bounds. Under different formalisms we can also see the POWERPLAY system as a way to increasingly augment the complexity of already explained problems \cite{Schmidhuber11POWERPLAY}. The approach from \cite{baranes2013active} can also be seen in this perspective where the space of policy parameters is explored in an increasing order of complexity.

One of the earliest works that tried to operationalize these concepts was made by \cite{Schmidhuber91curiosity}. More recently several researchers have extended the study to many other domains \cite{schmidhuber1995learning,Schmidhuber06,Singh2005intrinsically,Oudeyer2007}. Research in this field has considered new problems such as: situations where parts of the state space are unlearnable \cite{baranes2009riac,baranes2013active}; guide exploration in different spaces \cite{baranes2013active}; environmental changes \cite{Lopes12zeta}; empirical measures of learning progress \cite{Schmidhuber06,Oudeyer2007,baranes2009riac,baranes2013active,Hester13aamas,Lopes12zeta}; limited agents \cite{singh2010separating,sorg2010internal,Sequeira11IMmulti}; open-ended problems \cite{Singh2005intrinsically,Oudeyer2007}; autonomous discovery of good representations \cite{luciw2011artificial}; and selecting efficient exploration policies \cite{Lopes12ssp,Hester13aamas}. 

Some of these ideas are natural extensions to the active learning setting, e.g. time-variant problems, heteroscedastic domains but, usually due to limited formal understanding, theoretical results have been limited. Table \ref{tab:ActiveLearningVsIM} shows a comparison of the main qualitative differences between the traditional perspective and this more recent generalizations.

\begin{table*}[htbp]
	\centering
	\caption{Active Learning vs Artificial Curiosity}
		\begin{tabular}{c|c}
		  Active Learning & Artificial Curiosity \\\hline
			Learn with reduced time/data  & Learn with reduced time/data\\
			Fixed tasks										& Tasks change and are selected by the agent\\
			Learnable everywhere					& Parts might not be learnable\\
			Everything can be learned in the limit & Not everything can be learned during a lifetime\\
			Reduce uncertainty						& Improve progress\\
		\end{tabular}
	\label{tab:ActiveLearningVsIM}
\end{table*}

\subsection{Creating Representations}
\label{sec:CreatingRepresentations}

A very important aspect in any learning machine is to be able to create, or at least select, its own representations. In many cases (most?) the success of a learning algorithm is critically dependent on the selected representations. Any variant of feature selection is the most common approach for the problem and it is assumed that a large bank of features exist and the learning algorithm chooses a good sub-set of them, considering sparsity, or any other criteria. Nevertheless, the problem is not trivial and most heuristics are bound to fail in most cases \cite{guyon2003featureselection}.	

Some works focused just on the perceptual capabilities of agents. For instance, \cite{meng2008error} grows radial basis functions to learn mappings between sensory modalities by sampling locations with an high error. For the discussion on this document, particularly in this section, the most relevant works are those that not consider just what is the best representation for a particular task, but those that have a co-adaptation perspective and co-select the representation and the behavior. For instance \cite{ruesch2009evolving,schatz2009learning,rothkopf2009learning} study what is the relation between the behavior of an agent and the most representative retinal distribution.

Several works consider how to learn a good representations of the state-space of an agent while exploring an environment. These learned representations are not only good to classify regions but also to navigate and create hierarchies of behavior \cite{luciw2011artificial,bakker2004hierarchical}. Early works considered how a finite-automaton and an hierarchy could be learned from data \cite{pierce1995learning}.

Generalizations of those ideas consider how to detect regularities that identify non-static world objects and thus allowing to infer actions that change the world in the desired ways \cite{modayil2007autonomous}.

\subsection{Bounded Rationality}
\label{sec:BoundedRationality}

There are several models of artificial curiosity, or intrinsic motivation systems, that, in general, guide the behavior of the agent to novel situations. These models provide exploration bonuses, sometimes called intrinsic rewards, to focus attention on such novel situations. The advantages of such models for an autonomous agents are, in many situations, not clear.

An interesting perspective can be that of bounded rationality. Even if agents were able to see all the environment they might lack the reasoning and planning capabilities to behave optimally. Another way to see these works is to consider that the agent lives in a POMDP problem and, for some cases, it is possible to find a different reward function that mitigate some of the partial observability problem.

A very interesting perspective was approached with the definition of the \textit{optimal reward problem} \cite{sorg2010internal}. In here the authors consider that the learning agent is limited in its reasoning capabilities. If it tries to optimize the observed reward signal it will be sub-optimal in the task, and so another reward is found that allows the agent to learn the task. The authors have extended their initial approach to have a more practical algorithm using reward gradient \cite{sorg2010reward} and by comparing different search methods \cite{sorg2011optimal}. Recently the authors considered how the computational resources must be taken into account when choosing between optimizing a new reward or planning the next actions. Such search for an extra reward signal can also be used to improve coordination in a multi-agent scenario \cite{Sequeira11IMmulti}.

\subsection{Creating Skills}
\label{sec:OtherTopics}

When an animal is faced with a new environment there are an infinite number of different tasks that it might try to achieve, e.g. learn the properties of all objects or understand its own dynamics in this new environment. It can be argued that there is the single goal of survival and that any sub-division is an arbitrary construct. We agree with this view but we consider that such sub-division will create a set of reusable sub-goals that might provide advantages in the single main goal.

\begin{figure}
	\centering
		\includegraphics[width=1.0\columnwidth]{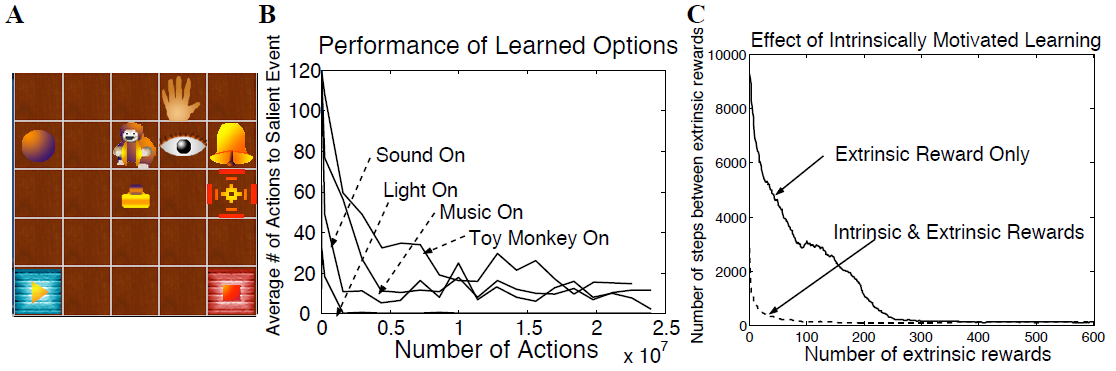}
	\caption{The playroom domain where a set of motor skills is incrementally created and learned resulting in a set of reusable, and hierarchical, repertoire of skills. (a) Playroom domain; (b) Speed of learning of various skills; (c) The effect of intrinsically motivated learning when extrinsic reward is present. From \cite{Singh2005intrinsically}.}
	\label{fig:playroom}
\end{figure}

This perspective on (sub) goal creation motivated one of the earliest computational models on intrinsic motivated systems \cite{barto2004intrinsically,Singh2005intrinsically}, see Figure \ref{fig:playroom}. There the authors, using the theory of options \cite{sutton1999between}, construct new goals (as options) every time the agent finds a new ''salient'' stimuli. In this toy example turning on a light, ringing a bell are considered reusable skills that might have an interest on latter stages and so if a skill is learned that reaches such state efficiently it will be able to learn complex hierarchical skills by combining the basic actions and the new learned skills.

The main criticism of those works is that the hierarchical nature of the problem was pre-designed and the saliency of novelty measures were tuned to the problem. To solve such limitations many authors have explored ways to autonomously define which skills much be created.  Next we will discuss different approaches that have been proposed to create new skills in various problems.

In regression problems several authors reduced the problem of learning a single complex task into learning a set of multiple simpler tasks. In problems modeled as MDPs authors have considered how to create macro-state or macro actions that can be reused in different problems of allow to create a complex hierarchical control system. After such division of a problem into a set of smaller problems it is necessary to decide what to learn at each time instant. For this, results from multi-armed bandits can be used, see \cite{Lopes12ssp} and Section \ref{sec:MAB}.

\subsubsection{Regression Models}

In problems that consist in learning forward and backward maps among spaces (e.g. to learn dynamical models of systems), authors have considered how to incrementally create a partition of the space into regions of consistent properties \cite{Oudeyer2007,baranes2009riac}. An initial theoretical study frames such model as a multi-armed bandits over a pre-defined hierarchical partition of the space \cite{maillard2012hierarchical}.

The set of skills that is created by the system might represent many different problems. Either an hierarchical decomposition of skills, but we can also see it as a decomposition of a problem in several, simpler, local problems. An example is the optimization setting of \cite{Krause08}. Here the authors try to find which regions of a given area must be sampled to provide more information about one of several environmental conditions. It considers an already known sub-division and learns the properties of each one. Yet, in real world applications, the repertoire of topics to choose from might not be provided initially or might evolve dynamically. 
The aforementioned works of \cite{Oudeyer2007,baranes2013active} consider initially a single region (a prediction task in the former and a control task in the latter) but then automatically and continuously constructs new region, by sub-dividing or joining previous existing ones. 

In order to discover affordances of objects and new ways to manipulate them, \cite{hart2008intrinsically} introduces an intrinsic reward that motivates the system to explore changes in the perceptual space. These changes are related to different motions of the objects upon contact from the robot arm.

A different perspective on regression methods is considering that the input space is a space of policy parameters and the output is whatever time-extended results of applying such policy. Taking into account this perspective, the approach from \cite{baranes2013active}, similarly to POWERPLAY \cite{Schmidhuber11POWERPLAY} and the approach from \cite{Auer11intrRL,lim2012autonomous}, explores the policy space in an increasing order of complexity of learning each behavior.

\subsubsection{MDP}

In the case of problems formulated as MDPs several researchers have defined automatic measures to create options or other equivalent state-action abstractions, see \cite{barto2003recent} for an early discussion. \cite{mannor2004dynamic} considered approaches such as online clustering of the state-action space using measures of connectivity, and variance of reward values. One such connectivity measure was introduced by \cite{mcgovern2001automatic} where states that are present in multiple paths to the goals are considered sub-goals and an option is initiated to reach them. These states can be seen as ''doors'' connecting between high-connected parts of the state-space. Other measures of connectivity have been suggested by \cite{menache2002q,simsek04icml,csimcsek2005identifying,simsek2008skill}. Even before the introduction of the options formalism, \cite{digney1998learning} introduced a method that would create skills based on reward gradients. \cite{hengst2002discovering} exploited the factored structure of the problem to create the hierarchy, by measuring which factors are more predictable and connecting that to the different levels of the hierarchy. A more recent approach models the problem as a dynamic bayesian network that explains the relation between different tasks \cite{jonsson2006causal}. 
Another perspective considers how to simultaneously learn different representations for the high-level and the lower level. By ensuring that neighbor states at the lower level are clustered in the higher level, it is possible to create efficient hierarchies of behavior \cite{bakker2004hierarchical}.

An alternative perspective on the creation of a set of reusable macro actions is to 
exploit commonalities in collections of policies \cite{thrun1995finding,pickett2002policyblocks}.

\subsection{Diversity and Competence}

For many learning problems we can define several spaces of parameters, usually the input parameters and the resulting behaviors are trivial concepts. Most of the previous concepts can be applied in different spaces and in many cases, and dependent on the metric of learning, there is a decision to be made on which of these spaces is better to use when guiding exploration. The robot might detect salient events in perceptual space, or generate new references, in the control space of a robot or on the environment space. Although coming from different perspectives: developmental robotics \cite{baranes2013active} and evolutionary development \cite{lehman2011noveltysearch} argue that exploration in the behavior space might be more efficient and relevant than in the space of the parameters that generate that behavior.

The first perspective proposed by \cite{lehman2011noveltysearch} is that many different genetic controller encodings might lead to very similar behaviors, and when considering also the morphological and environmental restrictions, the space of behaviors is much smaller than the space of controller encodings. The notion of diversity is not clear due to the redundancy in the control parameters, see \cite{mouret2011encouraging} for a discussion. It is interesting to note that in a more computational perspective, particle filters tend to also consider diversity criteria to detect convergence and improve efficiency \cite{gilks2002following}.

\begin{figure}
	\centering
		\includegraphics[width=0.7\columnwidth]{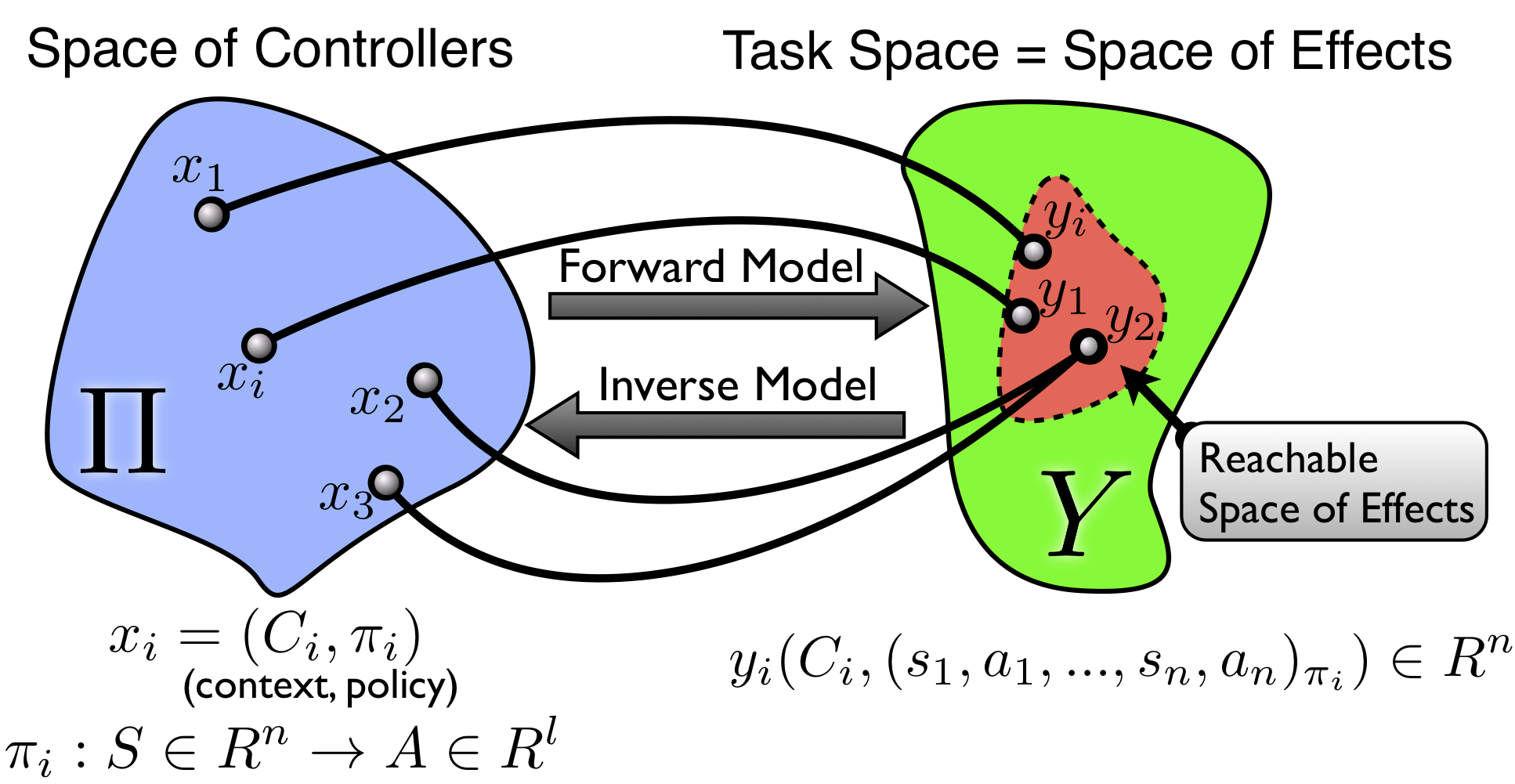}
	\caption{Model of the correspondences between a controller space and a task
space to be learned by a robot. Forward models deffine a knowledge of the
effects caused by the execution of a controller. Inverse models, which
deffine a skill or competence, are mechanisms that allow to retrieve one or
several controller(s) (if it exists) allowing to achieve a given effect (or
goal) yi in the task space.}
	\label{fig:spaces}
\end{figure}

From a robot controller point of view we can see a similar idea as proposed by \cite{baranes2010goalexplor}, see Figure \ref{fig:spaces}. In this case we consider the case of redundant robots where many different joint position lead to the same task space position of the robot. And so a dramatic reduction of the size of the exploration space is achieved. Also the authors introduced the concept of \textit{competence} where, and again for the case of redundant robots, the robot might prefer to be able to reach a larger volume of the task space, even without knowing all the possible solution to reach each point, than being able to use all the dexterity in a small part of the task space and not knowing how to reach the rest.

Other authors have considered also exploration in task space, e.g. \cite{Jamone11explor} and \cite{rolf2011online}. We can refer again to the works of \cite{Schmidhuber11POWERPLAY,lim2012autonomous} and see that they also consider as criteria having access to the more diversified set of policies possible.

\subsection{Development}
\label{sec:Development}

The previous discussion might lead us to think that a pure data-driven approach might be sufficient to address all the real world complexity. Several authors consider that data-driven approaches must be combined with pre-structured information. For examples artificial development considers that the learning process is guided not only by the environment and the data it is collect but also by the ''genetic information'' of the system \cite{elman1997rethinking,lungarella03develsurvey}.

In living organism, it is believed that maturational constraints help reduce the complexity of learning in early stages thus resulting in better and more efficient learning in the longer term. It does this by structuring the perceptual and motor space \cite{nagai2006learning,lee2007staged,macl07smc,lapeyre2011maturational,baranes2011interaction,OudeyerBaranesKaplan2013} or by developing intrinsic rewards that focus attention to informative experiences \cite{baldassarre2011intrinsic,Singh2010}, pre-dispositions to detect meaningful salient events, among many other aspects.

\subsection{Open Challenges}
\label{sec:futureCuriosity}

In a  broad perspective, open-ended learning and curiosity is still far from being a problem well understood, or even well formulated. Evolutionary models \cite{Singh2010} and recent studies in neurosciences \cite{Gottlieb13TICS} are starting to provide a more clear picture on if, and why, curiosity is an intrinsic drive in many animals. A clear understanding on why this drive exist, what triggers the drive to learn new tasks, and why agents seek complex situations will provide many insights on human cognition and on the development of autonomous and robust agents.

A related discussion is that a purely data-driven approach will not be able to consider such long-term learning problems. If we consider large domain problems, time-varying, the need for prior information that provide exploration constraints will be a fundamental aspect on any algorithm. This developmental constraints, and all genetic information, will be fundamental to any of such endeavor. We note that during learning and development it is required to co-develop representations, exploration strategies, learning methods, and hierarchical organization of behavior will require the introduction of novel theoretical frameworks.

\section{Interaction}
\label{sec:explorsocial}

\begin{figure}
	\centering
		\includegraphics[width=0.7\columnwidth]{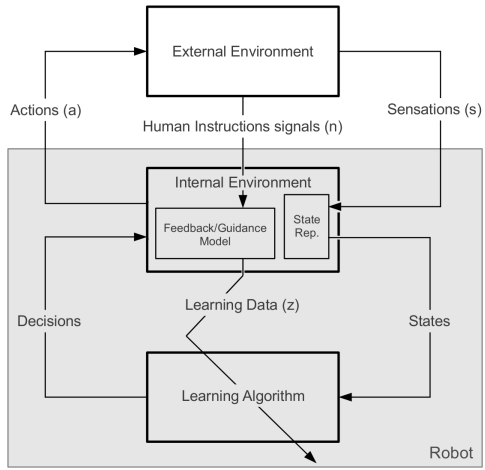}
	\caption{In many situations agents gather data from humans. These instructions need to be translated to a representation that is understood by the learning agent. From \cite{grizou2013robot}.}
	\label{fig:interlearn}
\end{figure}

\begin{figure}
	\centering
		\includegraphics[width=0.7\columnwidth]{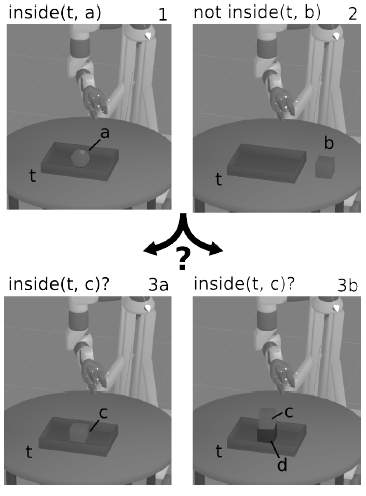}
	\caption{Active learning can also be used to instruct a robot how to label states allowing to achieve a common framing and providing symbolic representations that allow more efficient planning systems. In active learning of grounded relational symbols, the
robot generates situations in which it is uncertain about the symbol
grounding. After having seen the examples in (1) and (2), the
robot can decide whether it wants to see (3a) or (3b). An actively
learning robot takes its current knowledge into account and prefers
to see the more novel (3b). From \cite{Kulick13actsymbol}.}
	\label{fig:intersymbollearn}
\end{figure}

The previous sections considered active learning where the agents act, or make queries, and either the environment or an \textit{oracle} provides more data. Such abstract formalism might not be the best model when the oracle is a human with specific reasoning capabilities. Humans have a tremendous amount of prior knowledge, inference capabilities that allows them to solve very complex problems and so a benevolent teacher might guide exploration and provide information for learning. Feedback from a teacher takes the form of: initial condition for further self-exploration in robotics \cite{nicolescu2003natural}, information about the task solution \cite{Calinon07}, information about affordances \cite{ekvall2004interactive},  information about the task representation \cite{macl07affimit}, among others. Figure \ref{fig:interlearn} explains this process where the world state, the signals produced by the teacher and the signal required to the learning algorithms are not in the same representation and an explicit mechanism of translation is required. An active learning approach can also allow a robot to inquire a user about adequate state representations, see Fig.~\ref{fig:intersymbollearn}.

It has been suggested that \textit{interactive learning}, human-guided machine learning or learning with human in-the-loop,  might be a new perspective on robot learning that combines the ideas of learning by demonstration, learning by exploration, active learning and tutor feedback \cite{dillmann2000learning,dillmann2002interactive,fails2003interactive,nicolescu2003natural,Breazeal2004,lockerd2004tutelage,dillmann2004teaching}. Under this approach the teacher interacts with the robot and provides extra feedback. Approaches have considered extra reinforcement signals \cite{thomaz2008teachable}, action requests \cite{Grollman07dogged,macl09airl}, disambiguation among actions \cite{Chernova09jair}, preferences among states \cite{Mason2011}, iterations between practice and user feedback sessions \cite{judah2010rlandcritique,korupolu2012instructing} and choosing actions that maximize the user feedback \cite{knox2009interactively,Knox2010}. 

In this document we are more focused in active perspective and so it is the learner that has to ask for such information. Having a human on the loop we have to consider the cost in terms of tiredness of making many queries. Studies and algorithms have considered such aspect and addressed the problem of deciding when to ask. Most approaches will just ask to user whenever the information is needed \cite{nicolescu2001learning} or when there is high uncertainty \cite{Chernova09jair}. A more advanced situation considers making queries only when it is too risky to try experiments \cite{Doshi08actpomdp}. \cite{Cakmak2010desactlearn} compare the results when the robot has the option of asking or not the teacher for feedback and in a more recent work they study how can the robot make different types of queries including: label, features and demonstrations \cite{cakmak2011active,cakmak2012designing}.

Most of these systems have been developed to speed-up learning or to provide a more intuitive way to program robots. There are reasons to believe that an interactive perspective on learning from demonstration might lead to better results (even for the same amount of data). The theoretical aspects of these interactive systems have not been explored, besides the directly applied results from active learning. One justification for the need and expected gain of using such systems is discussed by \cite{Ross10effred}. Even if an agent learns from a good demonstration then, when executing that learned policy, its error will grow with $T^2$ (where $T$ is the horizon of the task). The reason being that any deviation from the correct policy moves the learner to a region where the policy has a worse fit. If a new demonstration is requested in that new region then the system learns not only how to execute a good policy but also how to correct from small mistakes. Such observation, as the authors refer, was already given by \cite{pomerleau1992neural} without a proof.

Another reason to use interactive systems is that when the users train the system they might become more comfortable with using it and accept it. See the work from \cite{ogata2003interactive} for a study on this subject. The queries of the robot will have the dual goal of allowing the robot to deal with its own limitations and give the user information about the robot's uncertainty on the task being learned \cite{fong2003robot,chao2010transparent}. 

There are many cases where the learning data comes directly from humans but no special uncertainty models are used. Such system either have an intuitive interface to provide information to the system during teleoperation \cite{kristensen1999interactive}, or it is the system that initiates questions based on perceptual saliency  \cite{lutkebohle2009curious}. There is also the case where the authors just follow the standard active learning setting (e.g. to learn a better gesture classification the system is able to ask the user to provide more examples of a given class \cite{francke2007real} even if for human-robot interfaces \cite{lee1996online}).

This section will start by presenting a perspective on the behavior of humans when they teach machines and the various ways in which a human can help a learning system. We then divide our review into systems for active learning from demonstration where the learner makes questions to the user and a second part where the teacher intervenes whenever it is required. Finally we show that sometimes it is important to try to learn explicit the teaching behavior of the teacher.

\subsection{Interactive Learning Scenarios}
\label{sec:InteractiveLearningScenarios}

The type of feedback/guidance that an human can provide depends on the task, the human knowledge, how easy it is to provide each type of information, the communication channels between the system and the user, among many other factors. For instance in a financial situation it is straightforward to attribute values to the outcomes of a policy but in some tasks, dancing for instance, it is much easier to provide trajectory information. In some tasks a combination of both is also required, for instance when teaching how to play tennis it is easy to provide a numeric evaluation of the different policies, but only by showing particular motions can a learner really improve its game. 

The presence of other agents in the environment creates diverse opportunities for different learning and exploration scenarios. We can view the other agents as teachers that can behave in different ways. They can provide: 
\begin{itemize}
	\item guidance on exploration
	\item examples
	\item task goals
	\item task solutions
	\item example trajectories
	\item quantitative or qualitative evaluation on behavior
	\item information about their preferences
\end{itemize}

By guiding exploration we consider that the agent is able to learn by itself but the extra feedback, or guidance, provided by the teacher will improve its learning speed. The teacher can be demonstrating new tasks and from this the learner might get several extra elements: the goal of the task, how to solve the task, or simply environment trajectories. Another perspective puts the teacher in a jury perspective of evaluating the behavior of the system, either providing directly an evaluation on the learner's behavior or by reveling his preferences. Several authors provided studies on how to model the different sources of information during social learning in artificial agents \cite{Noble02sociallearning,fmelo07unified,Nehaniv07ninebillion,macl09ab,Cakmak2010expsocpart,billing2010formalism}.

\begin{table*}
	\centering
	\caption{Interactive Learning Teachers}
	{\small
		\begin{tabular}{|l|l|}\hline
				Teacher 			& Examples \\\hline
				unaware  		& \cite{price03rlimitation} 					\\\hline
				batch					& 	\cite{Argall09lfdsurvey,lopes10imitationchapter,Calinon07}						\\\hline
				active 					&	Section \ref{sec:LearningByDemonstration}\\\hline
				teaching  		& \cite{cakmak2012designing,cakmak12algteach} \\\hline
				mixed					& \cite{katagami2000interactive,judah2010rlandcritique,thomaz2008teachable}\\\hline
				on-the-loop & \cite{Grollman07dogged,knox2009interactively,Mason2011}\\\hline
				ambiguous protocols & \cite{grizou2013robot}\\\hline
		\end{tabular}}
	\label{tab:InteractiveLearningTeachers}
\end{table*}
 
We can describe interactive learning system along another axis, and that is what type of participation the human has in the process. Table \ref{tab:InteractiveLearningTeachers} provides a non-exhaustive list of the different positions of a teacher during learning. First, the demonstrator can be completely \textit{unaware} that a learner is observing him and collecting data for learning. Many systems are like this and use the observation as a dataset to learn. Most interesting cases are those where the teacher is aware of the situation and provides the learner with a \textit{batch} of data; this is the more common setting. In the active approach the teacher is passive and only answers the questions of the learner (refer to Section \ref{sec:LearningByDemonstration}), while in the teaching setting it is the teacher that actively selects the best demonstration examples, taking into account the task and the learner's progress. Recent examples exist of human on-the-loop setting where the teacher observes the actions of the robot and only acts when it is required to make a correction or provide more data.

As usual all these approaches are not pure and many combine different perspectives. There are situations where different teachers are available to be observed and the learner chooses which one to observe \cite{price03rlimitation} where some of them might not even be cooperative \cite{Shon07imitmultdemons}, and even choose between looking at a demonstrator or just learn by self-exploration \cite{nguyen2011bootstrapping}.

\subsection{Human Behavior}
\label{sec:HumanBehavior}

Humans change the way they act when they are demonstrating actions to others \cite{nagai2009motionese}. This might help the learner by attracting attention to the relevant parts of the actions, but it also shows that humans will change the way a task is executed, see \cite{thomaz2009learning,kaochar2011towards,knox2012humans}.

It is clear now that when teaching robots there is also a change in behavior \cite{thomaz2006reinforcement,thomaz2008teachable,kaochar2011towards}. An important aspect is that, many times, the feedback is ambiguous and deviates from the mathematical interpretation of a reward or a sample from a policy. For instance, in the work of \cite{thomaz2008teachable} the teachers frequently gave a reward to exploratory actions even if the signal was used as a standard reward. Also, in some problems we can define an optimal teaching sequence but humans do not behave according to those strategies \cite{Cakmak2010opthumteach}.

\cite{kaochar2011towards} developed a GUI to observe the teaching patterns of humans when teaching an electronic learner to achieve a complex sequential task ( e.g. search and detect scenario ). The more interesting finding is that humans use all available channels of communication, including demonstration; examples; reinforcement; and testing. The use of testing varies a lot among users and without a fixed protocol many users will create very complex forms of interaction.


\subsection{Active Learning by Demonstration}
\label{sec:LearningByDemonstration}


Social learning, that is learning how to solve a task after seeing it being done has been suggested has an efficient way to program robots. Typically, the burden of selecting informative demonstrations has been completely on the side of the teacher. Active learning approaches endow the learner with the power to select which demonstrations the teacher should perform. Several criteria have been proposed: game theoretic approaches \cite{Shon07imitmultdemons}, entropy \cite{macl09airl,melo2010metric}, query by committee \cite{judah12active}, membership queries \cite{Melo13GBS}, maximum classifier uncertainty \cite{Chernova09jair}, expected myopic gain \cite{cohn2010selecting,cohn2011comparing} and risk minimization \cite{Doshi08actpomdp}.

One common goal is to find the correct behavior, defined as the one that matches the teacher, by repeatedly asking for the correct behavior in a given situation. Such idea as been applied in situations as different as navigation \cite{macl09airl,cohn2010selecting,cohn2011comparing,melo2010metric}, simulated car driving \cite{Chernova09jair} or object manipulation \cite{macl09airl}. 

\subsubsection{Learning Policies}
\label{sec:LearningPolicies}

Another learning task of interest is to acquire policies by querying an oracle. \cite{Chernova09jair} used support-vector machine classifiers to make queries to the teacher when it is uncertain about the action to execute as measured by the uncertainty of the classifier. They apply this uncertainty sampling perspective online, and thus only make queries in states that are actually encountered by the robot. A problem with this approach is that the information on the dynamics of the environment is not taken into account when learning the policy. To address this issue, \cite{melo2010metric} proposed a method that computes a kernel based on MDP metrics \cite{taylor2008bounding} that includes the information of the environment dynamics. In this way the topology of the dynamics is better preserved and thus better results can be obtained then with just a simple classifier with a naive kernel. They use the method proposed by \cite{montesano12actbetas} to make queries where there is lower confidence of the estimated policy.

Directly under the inverse reinforcement learning formalism, one of the first approaches were proposed by \cite{macl09airl}. After a set of demonstration it is possible to compute the posterior distribution of reward that explain the teacher behavior. By seeing each sample of the posterior distribution as a different expert, the authors took a query by committee perspective allowing the learner to ask the teacher what is the correct action in the state where there is higher disagreement among the experts (or more precisely where the predicted policies are more different). This work was latter extended by considering not just the uncertainty on the policy but the expected reduction in the global uncertainty \cite{cohn2010selecting,cohn2011comparing}.

The teacher can directly ask about the reward value at a given location \cite{regan2011eliciting} and it has been shown that reward queries can be combined with action queries \cite{Melo13GBS}.

The previous works on active inverse reinforcement learning can be seen as a way to infer the preferences of the teacher. This problem of \textit{preference elicitation} has been addressed in several domains \cite{furnkranz2010preference,chajewska2000making,braziunas2005local,viappiani2010optimal,Brochu07NIPS}.

\subsection{Online Feedback and Guidance}
\label{sec:OnlineCorrection}

Another approach is to consider that the robot is always executing and that a teacher/user might interrupt it at any time and assume the command of the robot. These corrections will act as new demonstrations to be incorporated in the learning process.

The TAMMER framework, and its extensions, considers how signals from humans can speed up exploration and learning in reinforcement learning tasks \cite{knox2009interactively,Knox2010}. The interesting aspect is that MDP reward is informational poor but it is sampled from the process while the human reinforcement is rich in information but might have stronger biases.  Knox \cite{knox2009interactively,Knox2010} presented the initial framework where the agent learns to predict the human feedback and then selects actions to maximize the expected reward from the human. After learning to predict such behavior during learning the agent will also observe the reward from the environment. The combination of both allows the robot to learn better using information given by the user will shape the reward function \cite{Ng99rewshap} improving the learning rate of the agent. Recently this process was improved to allow both processes to occur simultaneously \cite{knox2012reinforcement}.

It is important to take care to ensure that the shaping made by a human does not change the task. \cite{zhang2009policy} introduced a method were the teacher is able to provide extra rewards to change the behavior of the learner but, at the same time, considering that there is a limited budget on such extra rewards. Results showed that there are some tasks that are not possible to teach under a limited budget.

Other approaches considered that the learner can train by self-exploration and have several periods where the teacher is able to criticize its progress \cite{Manoonpong2010,judah2010rlandcritique}. 

Several work consider that initially the system will not show any initiative and will be operated by the user. Then as learning progresses the system will start acting according to the learned model while the teacher will act when a correction, or an exception, is needed. For instance, in the \textit{dogged learning} approach suggested in \cite{Grollman07dogged,grollman2007learning,grollman2008sparse} an AIBO robot is teleoperated and learns a policy from the user to dribble a ball towards a goal. After that training period the robot starts executing the learned policy but, at any time, the user has the possibility of resuming the teleoperation to provide eventual corrections. With this process a policy, encoded with a gaussian process, can be learned with better quality. A similar approach was followed in the work of \cite{Mason2011}. The main difference is that here the robot does not learn a policy and instead learns the preferences of the user and the interaction is done with a natural language interface. The authors consider a cleaning robot that is able to move objects in a room. Initially the robot as only a generic user profile that consider desired object locations, then after several interactions the robot moves the objects to the requested location. Every time the user says that the room is clean/tidy, the robot memorizes the configuration and through a kernel method is able to generalize what is a clean of not clean robot to different contexts.  With the advent of compliant robots the same approach can be made where the corrections are provided directly by moving the robot arm \cite{sauser2011iterative}.

An interesting aspect that was not explored much is to consider delays in the user's feedback. If such asynchronous behavior exist then the agent must decide how to act while waiting for the feedback \cite{cohn2012planning}.

\subsection{Ambiguous Protocols and Teacher Adaptation}
\label{sec:AmbiguousProtocols}

In most of the previous discussion we considered that the feedback signals provided by the teacher have a semantic meaning that is known to the learner. Nevertheless, in many cases the signals provided by the teacher might be too noisy or have unknown meaning. Several of these works fall under the learning from communication framework \cite{klingspor1997human}, where a shared understanding between the robot and the teacher is fundamental to allow good interactive learning sessions. 

The system in \cite{mohammad2010learning} automatically learns different interaction protocols for navigation tasks where the robot learns the actions it should make and which gestures correspond to those actions.  In \cite{macl11simul,grizou2013robot} the authors introduce a new algorithm for inverse reinforcement learning under multiple instructions with unknown symbols. At each step the learner executes an action and waits for the feedback from the user. This feedback can be understood as a correct/incorrect action, the name of the action itself or a silence. The main difficulty is that the user uses symbols that have an unknown correspondence with such feedback meanings. The learner assumes that the teacher feedback protocol and simultaneously estimates the reward function, the protocols being used and the meaning of the symbols used by the teacher. An early work consider such process in isolation and considered that learning the meaning of communication can be simplified by using the expectation from the already known task model \cite{kozima2001robot}.

Other works, such as \cite{lauria2002mobile,Kollar10ISER}, consider the case of learning new instructions and guidance signals for already known tasks, thus providing more efficient commands for instructing the robot. This algorithm is different from typical learning by demonstration systems because data is acquired in an interactive and online setting. It is different from previous learning by interaction systems in the sense that the feedback signals received are unknown.

The shared understanding between the teacher and the agents needs also to include a shared knowledge of the names of states. In \cite{Kulick13actsymbol} the authors take an active learning approach allowing the robot to learn state descriptions that are meaningful for the teacher, see Fig.~\ref{fig:intersymbollearn}.

\subsection{Open Challenges}
\label{sec:futureInteraction}

There are two big challenges in interactive systems. A first one is to clearly understand the theoretical properties of such systems. Empirical results seem to indicate that an interactive approach is more sample efficient than any specific approach taken in isolation. Another aspect is the relation between active learning and optimal teaching, where does not exist yet a clear understanding on the problems that can be learned efficiently but not taught and vice-versa.
The second challenge is to model accurately the human, or in general the cognitive/representational differences between the teacher and the learner, during the interactive learning process. This challenge include how to create a shared representation of the problem, how to create interaction protocols, and physical interfaces, that enables such shared understanding, and how to exploit the multi-modal cues that humans provides during instruction and interaction.

\section{Final Remarks}
\label{sec:FinalRemarks}

In this document we presented a general perspective on agents that, aiming at learning fast, look for the most important information required. To our knowledge it is the first time that a unifying look on methods and goals of different communities was made. Several further developments are still necessary in all these domains, but there is already the opportunity to a more multidisciplinary perspective that can give rise to more advanced methods.



\small
\bibliographystyle{apalike}
\bibliography{complete,macl}

~\\
\normalsize
\tableofcontents

\end{document}